\documentclass[11pt]{article}

\usepackage[preprint]{acl}

\usepackage{times}
\usepackage[table]{xcolor}

\usepackage{latexsym}
\usepackage{array}
\usepackage{algorithm}
\usepackage{algpseudocode}
\algrenewcommand\algorithmicrequire{\textnormal{Input:}}
\algrenewcommand\algorithmicensure{\textnormal{Output:}}
\algrenewtext{ElsIf}{\textnormal{else if}}
\algrenewcommand\algorithmicif{\textnormal{if}}
\algrenewcommand\algorithmicthen{\textnormal{then}}
\algrenewcommand\algorithmicelse{\textnormal{else}}
\algrenewcommand\algorithmicend{\textnormal{end}}

\usepackage{subcaption}
\usepackage{amsmath}
\usepackage{amssymb}
\usepackage{multirow}
\usepackage{float}
\usepackage[most]{tcolorbox}
\tcbset{
  thorprompt/.style={
    enhanced,
    colback=gray!6, colframe=black,
    boxrule=0.4pt, arc=2pt,
    left=6pt,right=6pt,top=6pt,bottom=6pt,
    titlerule=0pt
  }
}
\newtcolorbox{PromptBox}[1]{thorprompt, title=\textbf{#1}}

\tcbset{
  thorcase/.style={
    enhanced,
    colback=gray!6,
    colframe=black,
    boxrule=0.4pt,
    arc=2pt,
    left=6pt,right=6pt,top=6pt,bottom=6pt,
    before skip=6pt,
    after skip=6pt,
  }
}

\usepackage[T1]{fontenc}

\usepackage[utf8]{inputenc}

\usepackage{microtype}
\usepackage{booktabs}

\usepackage{inconsolata}

\usepackage{graphicx}
\usepackage{tabularx}
\title{\textbf{THOR}: A \underline{T}heta--Gamma \underline{H}ierarchical \underline{O}scillatory \underline{R}easoning Framework for Multi-hop QA}

\author{
  Ziyang Ling\textsuperscript{1,2}
  \and Ronald Xu\textsuperscript{1,2,*}
  \and Mingzhai Sun\textsuperscript{1,2,*}
  \\
  \textsuperscript{1}Suzhou Institute for Advanced Research, University of Science and Technology of China
  \\
  \textsuperscript{2}School of Biomedical Engineering, Division of Life Sciences and Medicine, \\
  University of Science and Technology of China, \\
  Hefei, Anhui, China
  \\
  \small{\textbf{Correspondence:} xux@ustc.edu.cn, mingzhai@ustc.edu.cn}
}

\begin{document}
\maketitle

\begin{abstract}
Multi-hop question answering requires retrieving and integrating evidence from multiple contexts. Despite the rapid progress of current research, multi-hop reasoning remains constrained by two persistent limitations: attention decay, where the model's focus on main question degrades as the reasoning chain grows, and error accumulation, where mistakes propagate across hops and compounds into final failure. Inspired by Theta--Gamma hierarchical oscillation which decouples global planning from local retrieval, enabling efficient attention transfer between hops and a verification and repair mechanism that interrupts the accumulation of errors in the wrong paths, we present \textbf{THOR}, a brain-inspired Theta--Gamma hierarchical oscillatory reasoning framework. Extensive comparative experiments and specific validation experiments on multi-hop QA benchmarks demonstrate that THOR improves answer accuracy and robustness while mitigating limitations, showcasing its generalization across different backbones. Our code is available at \url{https://github.com/ZaneLing/Theta-Gamma}.
\end{abstract}

\section{Introduction}
\label{sec:intro}
Facing questions with single-hop retrieval, current large language models (LLMs) have shown a powerful ability of reasoning in understanding question requirement, retrieving the supporting fact and generating a precise answer. Compared with single-hop retrieval question, multi-hop question answering is more challenging as it requires a model to determine a reasoning chain to integrate discrete facts from multiple passages and connect those facts with sequential reasoning to infer the final answer. Multi-hop question answering (QA) is a considerable ability to connect multiple pieces of information across documents which can be applied to many empirical domains as a fundamental tool.
\begin{figure}
    \centering
    \includegraphics[width=1\linewidth]{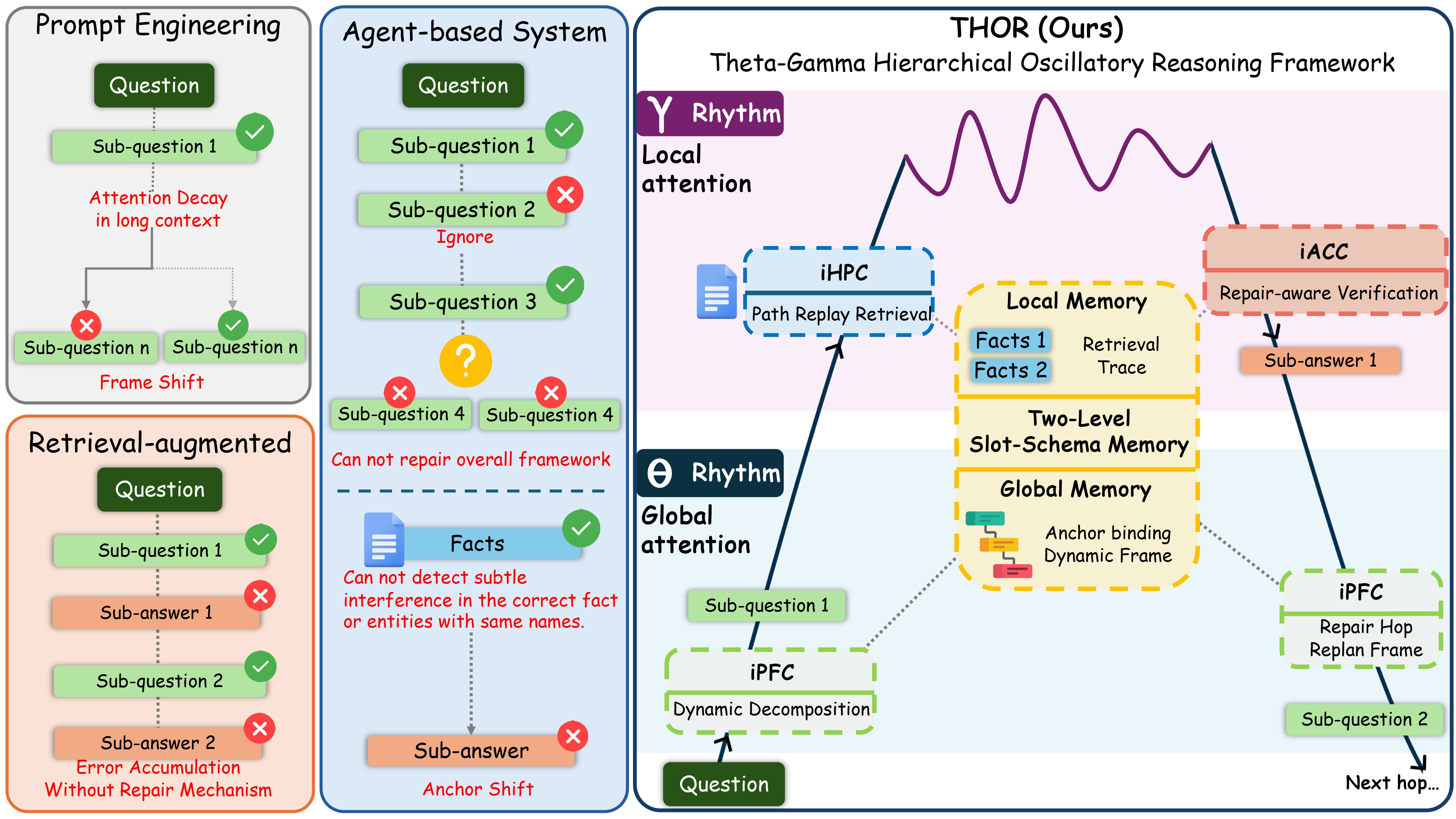}
    \caption{Comparison of multi-hop pipelines: Chain-of-Thought suffers attention decay as the reasoning chain becomes longer; Retrieval-augmented methods improve retrieval accuracy but face error accumulation problem without wrong path repair mechanism; Agent-based methods trigger repairing global frame when conflicts are detected but subtle errors such as anchor mismatch can not be found. In contrast, \textbf{THOR} decouples attention between global frame and local retrieval via a Theta--Gamma hierarchical oscillation and invoking both frame and anchor repair-aware backtracking mechanism.}
    \label{fig:introfig}
\end{figure}
With the growing progress of LLMs and reasoning-oriented architectures, current methods have achieved novel innovations across multiple dimensions and levels \citep{mavi2024multihop, plaat2025multistep}, demonstrating visible improvements in effectiveness. We conducted extensive testing on various methods and specifically analyzed numerous error cases. We discovered that these error cases can be attributed to two deep-seated reasons. One major limitation is attention decay, referring to the progressive drift of model focus as the reasoning chain lengthens. Another key bottleneck is error accumulation, which denotes the complete collapse of the reasoning path caused by subtle errors because of lack of error perception and correction. Together, these issues lead to increased hallucinations, unstable reasoning, and reduced robustness in future complicated reasoning.\\
Motivated by these limitations, clear experiments ~\citep{su2025multihopquestionansweringhumans} have demonstrated that human accuracy significantly outperforms all AI methods. What interests us most is the reasoning mechanism of the human brain when meeting multi-hop questions and how to resolve the two limitations above. By distinguishing the most significant differences, we find that human reasoning is a complex holistic structure, coordinated through neural oscillations with different brain regions. A recurring theme is Theta--Gamma neural hierarchical oscillation mechanism that $\theta$ rhythm provides a slower temporal scaffold that organizes and prioritizes goal-relevant processing, while $\gamma$ rhythm supports faster local computations~\citep{lakatos2008entrainment, lundqvist2016gamma}. Collaboration between multiple brain regions is controlled by Theta-Gamma neural hierarchical oscillations, each of which fulfills a distinct role. Together, these findings motivate a brain-inspired view of multi-hop reasoning as a controlled, hierarchical, and coordinated process to mitigate attention decay and error accumulation.\\
Inspired by the human brain, we introduce \textbf{THOR}, a Theta--Gamma hierarchical oscillatory and repair-aware reasoning framework for multi-hop QA. We make three key contributions:
(1) We propose a multi-hop reasoning framework with a logic loop controlled by oscillating , featuring an awareness of errors and an error-correction mechanism.
(2) THOR explicitly targets and mitigates two fundamental limitations in long reasoning chains multi-hop reasoning.
(3) We conduct extensive experiments of comprehensive perspectives to demonstrate accuracy, effectiveness and generalization.
\section{Related Work}
\label{sec:rl}
Multi-hop question answering (QA) requires integrating evidence across multiple contexts, and recent work has improved LLM-based multi-hop QA along several representative directions. First, prompt-engineering methods encourage stepwise reasoning through intermediate facts. Typical Chain-of-Thought (CoT)~\citep{wei2023chainofthoughtpromptingelicitsreasoning} prompting elicits multi-step derivations by explicitly generating reasoning traces, which can improve compositional reasoning. With the progress of retrieval-augmented generation, retrieval optimization methods target the evidence acquisition stage, aiming to provide more complete and relevant contexts for downstream reasoning. Chain-of-RAG (CoRAG)~\citep{wang2025chainofretrievalaugmentedgeneration} iteratively performs retrieval conditioned on intermediate reasoning states, enabling multi-hop evidence accumulation beyond one-shot retrieval. As the agentic system gains popularity, agent-based methods treat LLMs as decision-making agents that can decompose tasks, verify intermediate steps, and revise reasoning traces, such as Tree-Of-Reviews~\citep{jiapeng2024treereviewstreebaseddynamic} and ReAgent~\citep{zhao2025reagentreversiblemultiagentreasoning} which introduces reversible multi-agent reasoning with structured review and backtracking to mitigate wrong-path reasoning. While all previous works focus on a specific multi-hop QA method, our approach targets the pipeline inspired by human brain.

\begin{figure}
    \centering
    \includegraphics[width=1\linewidth]{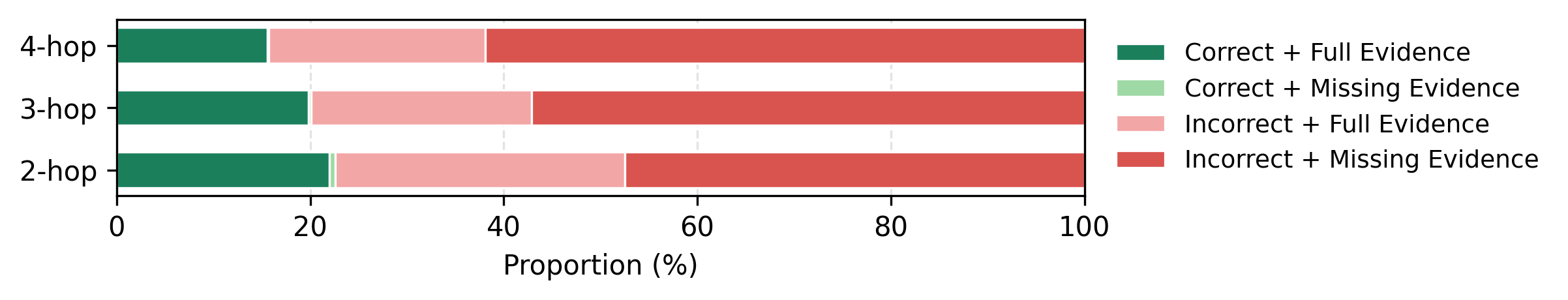}
    \includegraphics[width=1\linewidth]{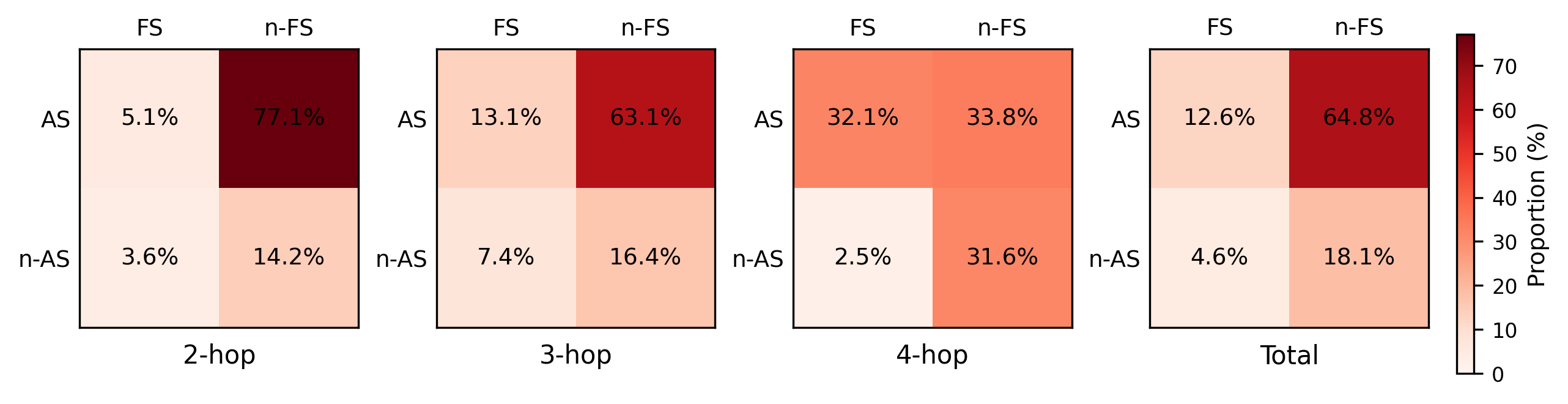}
    \caption{Current methods still have the problem of attention decay. Frame shift (FS) and anchor shift (AS) caused by attention decay account a significant proportion in the erroneous cases.}
    \label{fig:limit}
\end{figure}

\begin{figure*}
    \centering
    \includegraphics[width=1\linewidth]{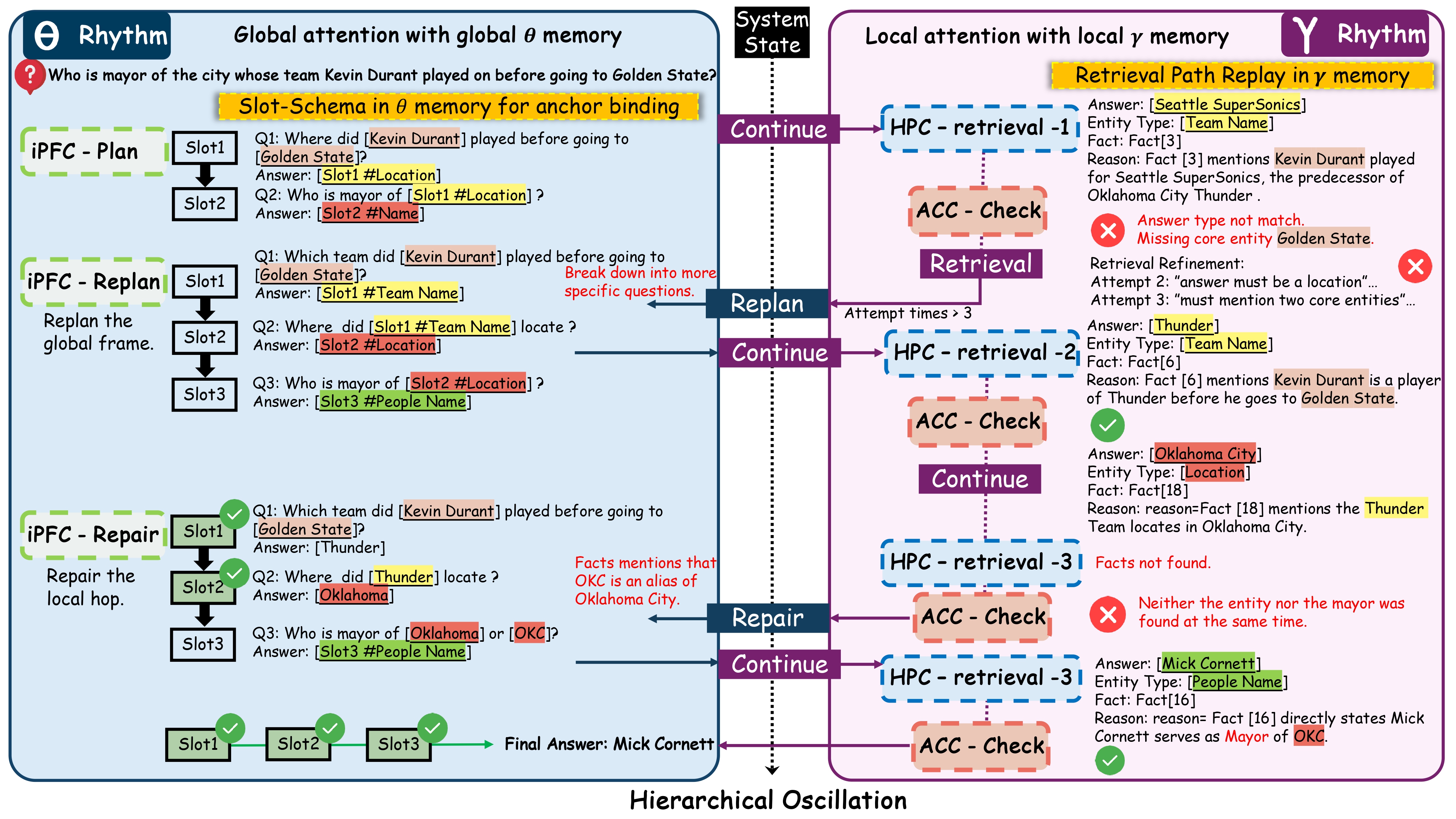}
    \caption{Overview of the framework of THOR. THOR alternates between a global $\theta$ rhythm that maintains attention on the reasoning frame, and a local $\gamma$ rhythm that executes each hop via focused retrieval and verification. Attention decay is reduced by hierarchical oscillatory attention transfer which is controlled by a discrete system state decided by explicit signals. Error accumulation is suppressed by a slot-schema based memory and per-hop checks in $\gamma$ rhythm which trigger targeted backtracking strategy.}
    \label{fig:frame}
\end{figure*}

\section{Limitations of Previous Methods}
\label{sec:limitations}
Multi-hop reasoning becomes increasingly brittle as hop depth grows (Fig.~\ref{fig:limit}), exhibiting two recurring drift phenomena. Frame shift arises when intermediate steps deviate from the intended decomposition frame, causing off-target sub-questions and irrelevant evidence retrieval. Even when the global frame remains plausible, anchor shift can occur at the entity level: the reasoning detaches from the correct anchor entity, and the retrieved evidence no longer grounds the intended sub-question. We attribute both phenomena to attention decay, where model focus progressively degrades over long reasoning chains, weakening constraint tracking over both the global frame and entity bindings.
Methods with repair mechanism such as ReAgent~\citep{zhao2025reagentreversiblemultiagentreasoning} perform backtracking and repair for the current hop but lack the ability to repair the overall frame. Once such drift occurs, errors in the early hop can persist and accumulate through hops which can not be detected, and thus multi-hop reasoning demands explicit verification and repair mechanisms beyond retrieval alone. 

\section{Methodology}
\subsection{Theta--Gamma Hierarchical Oscillation in Human Brain}
\label{sec:theta-gamma}
Human brain constructs a hierarchical temporal regulatory framework through cross-frequency coupling between $\theta$ waves and $\gamma$ waves, namely Theta--Gamma hierarchical oscillation, to achieve dynamic transfer of attention between global and local, as well as effective suppression of error accumulation. $\theta$ oscillations act as a master clock, providing periodic reset signals that refresh attentional resources at the end of each reasoning hop and preventing resource exhaustion. In $\gamma$ rhythm, its amplitude varies dynamically across the $\theta$ rhythm. The rising edge facilitates high-intensity encoding, the peak region filters redundant features, and the falling edge focuses on error checking and result integration. Simultaneously, dynamic resource scheduling mechanism mediated by the cholinergic system ensures the brain achieves spatial scheduling. High acetylcholine levels stabilize $\gamma$ rhythm for active computation, while low levels modulates $\theta$ rhythm to facilitate cross-step information transmission~\citep{FiebelkornKastner2019,LismanJensen2013,FisahnEtAl1998,Buzsaki2002}. To ensure reasoning accuracy across multiple steps, Theta--Gamma hierarchical oscillation implements effective mechanisms with the coordination among multiple brain regions. The Prefrontal Cortex sends predictive signals via $\theta$ rhythm to be compared with real-time hippocampal $\gamma$ encoding. Mismatches trigger theta phase shifts and dopamine release to initiate immediate correction. A computation--verification--adjustment circuit feedback between the PFC, hippocampus, and anterior cingulate cortex operate within each Theta--Gamma Oscillation, ensuring errors are corrected before they enter subsequent reasoning stages. Additionally, information is stored as $\theta$ phase-locked $\gamma$ sequences. These ordered memory traces provide a structured basis for the brain to systematically retrace and correct accumulated errors after the reasoning process is complete~\citep{LismanJensen2013,JonesWilson2005,SchultzDayanMontague1997,FosterWilson2007}.

\subsection{THOR: Hierarchical Oscillatory Reasoning Framework}
\label{subsec:overview}
Inspired by these findings, we introduce THOR, a Theta--Gamma hierarchical oscillatory reasoning framework for multi-hop reasoning. THOR organizes multi-hop reasoning as a closed-loop process controlled by Theta--Gamma hierarchical oscillations as shown in Fig.~\ref{fig:frame} over a two-level slot-schema based memory. At each hop, the framework plans the next reasoning step under the system states, retrieves and integrates evidence and verifies faithfulness and consistency, and triggers targeted repair.
\subsubsection{Global Theta Rhythm}
\label{sec:app_ipfc}
Operating as a slow master clock, global $\theta$ rhythm maintains attention on the global $\theta$ memory according to the main question to dynamically adjust the multi-hop reasoning frame under explicit constraints, enabling controlled repair/replan that prevents deep-chain error propagation. We represent the reasoning frame in global $\theta$ memory as a slot-schema composed of slots with corresponding entity types. Each hop binds entity-centered slot by writing the correct answer according to the evidence, which explicitly enforces entity binding and prevents drifting to irrelevant entities. Only when all required slots are filled can the final answer be deterministically composed from the completed schema.
\begin{tcolorbox}[
  colback=white,
  colframe=black,
  boxrule=0.9pt,
  arc=1.2mm,
  left=1.6mm,right=1.6mm,top=0.8mm,bottom=0.2mm,
  width=\linewidth
]
\small
\textbf{\texttt{===== Global Theta Memory}} $\mathcal{M}^{\theta}$ \textbf{=====}\\
\hspace*{2mm}Main question: $Q$ \\
\hspace*{2mm}Sub-questions: $sq_{1:t}$ \quad \\
\hspace*{2mm}Core entities: $ce_{1:t}$ \quad \\
\hspace*{2mm}Expected answer types: $\tau_{1:t}$ \quad \\
\hspace*{2mm}Execution state (global slots):\\
\hspace*{4mm}Current expected sub-answer: $sa_{1:t}$\\
\hspace*{4mm}Completion flag: $ok_{1:t}$
\end{tcolorbox}
In global $\theta$ rhythm, PreFrontal Cortex-inspired module (iPFC) performs global frame adjustment by reading and writing the global $\theta$ memory:
\[
\mathcal{M}^{\theta'}:(sq_t,ce_t,\tau_t)\leftarrow \text{iPFC}\big(Q,\mathcal{M}^{\theta}\big)
\]
There are two repair modes according to the two system state control: (1) Repair: When the system state enters repair state, iPFC preserves the validated prefix hops and revises only the current hop specification using refinement feedback from $\gamma$ rhythm. This operation performs targeted edits such as alias expansion or micro-decomposition; (2) Replan: When failures persist and the per-hop retry budget is exhausted (default $I_{\max}=3$), the controller escalates to replan state, forcing iPFC to backtrack and update the global reasoning frame to avoid unbounded local loops. Unlike repair state, replan state is not restricted to the current hop, it may revise earlier bridge steps or regenerate a more coherent decomposition consistent with the main question constraints.

\subsubsection{Local Gamma Rhythm}
As a fast $\gamma$ execution clock, hippocampus-inspired module (iHPC) and anterior cingulate cortex-inspired module (iACC) operates over the local $\mathcal{M}^{\gamma}$ to complete a bounded single-hop by retrieving, integrating, and verification-gating evidence, producing actionable signals before committing updates back to the global $\theta$ frame. 
\begin{tcolorbox}[
  colback=white,
  colframe=black,
  boxrule=0.9pt,
  arc=1.2mm,
  left=1.6mm,right=1.6mm,top=0.8mm,bottom=0.2mm,
  width=\linewidth
]
\small
\textbf{\texttt{===== Local Gamma Memory}} $\mathcal{M}^{\gamma}$ \textbf{=====}\\
\hspace*{2mm}Hop $t$ trace at attempt $i$:\\
\hspace*{4mm}Retrieved evidence: $e_i$ \\
\hspace*{4mm}Predicted sub-answer: $\hat{sa}_i$ \\ 
\hspace*{4mm}Predicted type: $\hat{\tau}_i$ \\
\hspace*{4mm}Predicted reason: $r_i$ \\
\hspace*{4mm}Refinement: $rf_{i-1}$ \\
\hspace*{4mm}Verifier signal: $Acc_i=\langle \text{C}_{e_i},\text{C}_{\hat{\tau}_i},\text{C}_{r_i}, rf_i\rangle$
\end{tcolorbox}

In $\gamma$ rhythm, iHPC performs topic and context aware strategy to retrieve more accurate evidence with iterative refinements from iACC, as shown in Algorithm~\ref{alg:ihpcc-retrieval}. In particular, $\rho_i$ is the optimization guidance obtained by iHPC through replaying the paths of all retrieval attempts in this hop, while $rf_i$ is the optimization of retrieval achieved by iACC by mining the unmatched evidence content of a single attempt.
\begin{algorithm}[htb]
\caption{iHPC: Topic/Context-aware Evidence Retrieval}
\label{alg:ihpcc-retrieval}
\begin{algorithmic}[1]
\Require $ce_t,\ sq_t,\ rf_{i-1},\ \rho_{i-1},\ i,\ t$
\Ensure $e_i,\ E^{\text{topic}}_{t,i},\ E^{\text{ctx}}_{t,i}$
\Statex Enc: BGE-M3 \quad Rerank: BGE-Reranker
\State $\tilde{ce}_{t,i}\leftarrow \mathrm{Refine}(ce_t;\ rf_{i-1},\rho_{i-1})$
\State $u \leftarrow \mathrm{Enc}(\tilde{ce}_{t,i})$
\Statex (1) Topic-top-3:
\State $s_{\text{topic}}(z)\!=\!\cos\!\big(u,\mathrm{Enc}(z)\big),\ z\in\mathcal{T}$
\State $E^{\text{topic}}_{t,i}\leftarrow \mathrm{TopK}_{3}\big(\mathcal{T},\ s_{\text{topic}}\big)$
\State $\delta_i \leftarrow \mathbb{I}\!\big[\mathrm{Insuff}(E^{\text{topic}}_{t,i})\big]$ \Comment{iACC-verified}
\Statex (2) Context-top-3 (fallback):
\State $s_{\text{ctx}}(f)\!=\!\cos\!\big(u,\mathrm{Enc}(\mathrm{ctx}(f))\big),\ f\in\mathcal{F}$
\If{$\delta_i=1$}
  \State $E^{\text{ctx}}_{t,i}\leftarrow \mathrm{TopK}_{3}\big(\mathcal{F},\ s_{\text{ctx}}\big)$
\Else
  \State $E^{\text{ctx}}_{t,i}\leftarrow \emptyset$
\EndIf
\State $e_i \leftarrow$ $E_{t,i}\leftarrow E^{\text{topic}}_{t,i}\cup E^{\text{ctx}}_{t,i}$
\end{algorithmic}
\end{algorithm}

When the iHPC retrieves and provides the answer sequence, iACC performs repair-aware verification in Algorithm~\ref{alg:check-acc} to check three kinds of mismatch and emit structured feedback for the controller to trigger retrieval refinement or plan repair.
\begin{algorithm}[htb]
\caption{iACC: Repair-Aware Verification}
\label{alg:check-acc}
\begin{algorithmic}[1]
\Require $sq_t,\ e_i,\ ce_t,\ \hat{\tau}_i,\ \tau_t,\ \hat{sa}_i,\ r_i$
\Ensure $Acc_i=\langle C_{e_i},C_{\hat{\tau}_i},C_{r_i}\rangle,\ C_{\cdot}\in\{0,1\}$
\Statex Verifier: iACC (LLM)
\Statex (1)Evidence Anchoring Check
\State $C_{e_i}\leftarrow \mathbb{I}\!\big[\mathrm{Mention}(ce_t,\ e_i)\big]$
\Statex \Comment{normalize/tokenize + string match}
\Statex (2)Type Alignment Check
\State $C_{\hat{\tau}_i}\leftarrow \mathbb{I}\!\big[(sq_t,\hat{sa}_i)\ \Rightarrow\ \tau_t\big]$
\Statex \Comment{LLM judges whether $\hat{sa}_i$ match $\tau_t$}
\Statex (3)Evidence--Answer Support
\State $C_{r_i}\leftarrow \mathbb{I}\!\big[(sq_t,e_i,\hat{sa}_i,r_i)=\textsc{Support}\big]$
\Statex \Comment{LLM verifies $e_i$ supports $\hat{sa}_i$ under $sq_t$}
\State $Acc_i\leftarrow \langle C_{e_i},C_{\hat{\tau}_i},C_{r_i}\rangle$
\end{algorithmic}
\end{algorithm}

\subsubsection{Theta--Gamma Oscillatory Controller}
\label{sec:tg_controller}
Rhythm oscillation between $\theta$ rhythm and $\gamma$ rhythm is governed by a discrete system state selected at the end of each hop, including continue, retrieve, repair and replan, which determines whether the controller stays in local $\gamma$ rhythm or switches back to global $\theta$ rhythm. The controller maps $Acc_i$, the retry counter $i$ and max retry $I_{max}$ to a discrete system state control $c_t$ to control rhythm oscillation in Algorithm~\ref{alg:tg-switch}.
\begin{algorithm}[htb]
\caption{Rhythm Oscillation via System State}
\label{alg:tg-switch}
\begin{algorithmic}
\Require $Acc_i=\langle C_{e_i},C_{\hat{\tau}_i},C_{r_i}\rangle$, retry $i$, budget $I_{\max}$, $\mathcal{M}^{\theta}$
\Ensure $c_t$
\If{$Acc_i=\langle 1,1,1\rangle$}
  \State $c_t \leftarrow \textsc{Continue}$ \hfill $(\theta\!\rightarrow\!\gamma)$
  \State $\triangleright$ commit; write $\rightarrow \mathcal{M}^{\theta}$; next-hop $\gamma$
\ElsIf{ $C_{e_i}=0$}
  \State $c_t \leftarrow \textsc{Retrieve}$ \hfill $(\gamma\!\rightarrow\!\gamma)$
  \State $\triangleright$ refine query/evidence
\ElsIf{ $i>I_{\max}$}
  \State $c_t \leftarrow \textsc{Replan}$ \hfill $(\gamma\!\rightarrow\!\theta)$
  \State $\triangleright$ escalate; revise global frame
\Else
  \State $c_t \leftarrow \textsc{Repair}$ \hfill $(\gamma\!\rightarrow\!\theta\!\rightarrow\!\gamma)$
  \State $\triangleright$ rewrite hop spec; resume $\gamma$
\EndIf
\end{algorithmic}
\end{algorithm}

\section{Experiments}
\subsection{Experimental Setup}
\paragraph{Dataset.}
\label{sec:setup_datasets}
Three multi-hop QA benchmarks were used: (1)HotpotQA ~\citep{yang2018hotpotqadatasetdiverseexplainable} contains 113k questions; (2)2WikiMultiHopQA ~\citep{ho-etal-2020-2wikimultihopqa} is a large-scale dataset explicitly designed for cross-document reasoning; (3)MuSiQue ~\citep{trivedi2022musiquemultihopquestionssinglehop} focuses on compositional reasoning.
\paragraph{Metrics.}
\label{sec:setup_metrics}
We use Exact Match (EM) and F1 scores for the QA evaluation. In order to further verify the role of the framework in alleviating constraints, we further defined Frame Shift Rate and Anchor Shift Rate two metrics: (1) Frame Shift Rate (FSR). We use a \emph{fixed} LLM judge GPT-4o with temperature $=0$ to evaluate frame shift. Define step-level frame alignment ${fa}_t(x)\in\{0,1\}$, where ${fa}_t(x)=1$ if the predicted decomposition matches the gold decomposition. Then the Frame Shift Rate (FSR) is the proportion of off-frame steps among all predicted steps:
\[
\text{FSR}
=\frac{\sum_{x}\sum_{t=1}^{h}\bigl(1-{fa}_t(x)\bigr)}
{\sum_{x} h}.
\]
(2) Anchor Shift Rate (ASR). We use a lightweight anchor judge to evaluate anchor shift. Define hop-level anchor alignment ${aa}_t(x)\in\{0,1\}$, where ${aa}_t(x)=1$ if the anchor entity in the predicted sub-question is mentioned in the retrieved evidence at hop $t$. Then the Anchor Shift Rate (ASR) is the proportion of anchor-missing hops among all executed hops:
\[
\text{ASR}
=\frac{\sum_{x}\sum_{t=1}^{h}\bigl(1-{aa}_t(x)\bigr)}
{\sum_{x} h}.
\]
All metrics details present in Sec~\ref{sec:app_metrics}.
\paragraph{Baselines.}
We compared our methods with recent works of several directions: (1)Prompt-engineering methods such as typical Chain-of-Thought (CoT)~\citep{wei2023chainofthoughtpromptingelicitsreasoning}, Tree-of-Thought (ToT)~\citep{yao2023treethoughtsdeliberateproblem}, Self-prompted CoT (SP-CoT)~\cite{wang-etal-2023-self-prompted}, FSM~\citep{wang2024fsmfinitestatemachine} and Least-to-Most~\citep{zhou2023leasttomostpromptingenablescomplex}; (2)Retrieval optimization methods make efforts on retrieval, including Single-step~\citep{izacard2022atlasfewshotlearningretrieval}, Self-Ask~\citep{press-etal-2023-measuring}, IRCoT~\citep{trivedi-etal-2023-interleaving},  RetGen~\citep{shao2023enhancingretrievalaugmentedlargelanguage}, by chain-of-RAG (CoRAG)~\citep{wang2025chainofretrievalaugmentedgeneration}, EfficientRAG~\citep{zhuang2024efficientragefficientretrievermultihop}, ComposeRAG~\citep{wu2025composeragmodularcomposablerag}, FLARE~\citep{jiang-etal-2023-active}, ProbTree~\citep{cao-etal-2023-probabilistic}, HippoRAG~\citep{gutiérrez2025hipporagneurobiologicallyinspiredlongterm} and BeamAggR~\citep{chu2024beamaggrbeamaggregationreasoning}.
(3)Agent-based methods treat the LLMs as agents, such as PRISM~\cite{nahid-rafiei-2025-prism}, Chain-of-Agents~\citep{zhang2024chainagentslargelanguage}, GEAR~\citep{shen2025geargraphenhancedagentretrievalaugmented}, Search-o1~\citep{li2025searcho1agenticsearchenhancedlarge}, Tree-Of-Reviews (ToR)~\citep{jiapeng2024treereviewstreebaseddynamic}, KAG~\citep{liang2024kagboostingllmsprofessional} and specific multi-agent system like ReAgent~\citep{zhao2025reagentreversiblemultiagentreasoning}, RopMura~\citep{wu2025talkrightspecialistsrouting} and BELLE~\citep{zhang2025bellebilevelmultiagentreasoning}.
\paragraph{Implementation.} We use GPT-3.5-turbo as the backbone of THOR for our experiments, setting the maximum length context window to 4096. In the main experiment, we set max attempt retry parameter $I_{max}$ to 3. In the ablation study, we replaced iHPC with bm25~\citep{robertson1994simple} retrieval and replaced iPFC with single LLM without repair mechanism. We use BGE~\citep{chen2024m3embedding} to improve evidence retrieval.

\subsection{Main Results}
As shown in Table~\ref{tab:main_results}, THOR outperforms the vast major methods and particularly, THOR$^\dagger$ with GPT-4o achieves the best EM and F1 across all datasets. On the most challenging MuSiQue dataset, we achieved the best F1 score of 52.1 simply by using gpt-3.5-turbo. Prompt engineering methods does not have any special optimizations, so the improvement is limited especially in MuSiQue; Compared with prompt-engineering methods, retrieval-augmented methods significantly improves by retrieving more accurate evidence; Agent-based frameworks can repair errors by detecting conflicts but without explicit global control and precise error location.
\begin{table}[t]
\centering
\small
\setlength{\tabcolsep}{4pt}
\begin{tabular}{l|cc|cc|cc}
\toprule
Dataset & 
\multicolumn{2}{c|}{HotpotQA} &
\multicolumn{2}{c|}{2WikiQA} &
\multicolumn{2}{c}{MuSiQue} \\
Metrics &
EM & F1 & 
EM & F1 &
EM & F1 \\
\midrule
\multicolumn{7}{l}{\cellcolor{gray!15}\textbf{Prompt Engineering Methods}} \\
\midrule
CoT     & 40.5 & 46.5 & 36.2 & 42.3  & 21.1 & 24.9 \\
SP-CoT  & 33.2 & 42.9 & 30.1 & 34.7  & 23.7 & 25.7 \\
FSM     & 33.1 & 46.0 & 36.1 & 49.3  & 22.2 & 26.2 \\
ToT     & 36.9 & 43.0 & 40.1 & 48.4  & 19.2 & 22.2 \\
\midrule
\multicolumn{7}{l}{\cellcolor{gray!15}\textbf{Retrieval-augmented Methods}} \\
\midrule
Single-step  & 48.7 & 55.3 & 38.1 & 42.9 & 14.1 & 16.5 \\
Self-Ask     & 44.5 & 49.4 & 40.5 & 46.9 & 13.4 & 16.7  \\
IRCoT        & 51.2 & 56.2 & 50.7 & 56.8 & 23.1 & 25.2  \\
Iter-RetGen  & 45.9 & 61.1 & 36.0 & 48.1 & 26.4 & 42.0  \\
FLARE        & 50.8 & 56.1 & 58.2 & 60.1 & 31.1 & 32.2  \\
ProbTree     & 56.3 & 60.4 & 64.3 & 67.9 & 30.2 & 33.5  \\
EfficientRAG & 52.9 & 57.9 & 47.7 & 51.6 & 24.7 & 26.5  \\
BeamAggR     & 55.6 & 62.9 & 66.1 & 71.6 & 36.7 & 39.0  \\
ComposeRAG   & 55.8 & 70.2 & 72.8 & 74.0 & 32.8 & 37.6  \\
CoRAG        & 56.3 & 69.8 & 72.5 & 77.3 & 30.9 & 42.4  \\
\midrule
\multicolumn{7}{l}{\cellcolor{gray!15}\textbf{Agent-based Framework Reasoning}} \\
\midrule
CoA       & 39.1 & 55.8 & 57.5 & 69.7 & 23.9 & 36.1 \\
HippoRAG  & 52.8 & 71.7 & 63.3 & 72.5 & 35.3 & 51.7 \\
GEAR      & 50.4 & 54.6 & 47.4 & 52.3 & 25.6 & 27.3 \\
ToR       & 38.2 & 50.4 & 29.0 & 37.0 & 13.2 & 22.1 \\
PRISM     & 54.2 & 67.0 & 48.6 & 57.0 & 31.2 & 41.8 \\
RopMura   & 49.2 & 53.1 & 58.8 & 63.2 & 29.9 & 31.7 \\
KAG       & 60.3 & 78.2 & 68.1 & 78.1 & 34.8 & 48.9 \\
ReAgent  & 63.0 & \underline{79.5} & 71.1 & \underline{79.3} & 37.1 & 51.5 \\
Search-o1  & 45.2 & 57.3 & 58.0 & 71.4 & 16.6 & 28.2 \\
BELLE     & 59.2 & 66.5 & 69.7 & 75.7 & \underline{50.5} & 42.1 \\
\midrule
\textbf{THOR}    & \underline{69.3} & 76.1 & \underline{75.6} & 78.6 & 48.5 & \underline{52.1} \\
\textbf{THOR}$^\dagger$   & \textbf{72.1} & \textbf{81.4} & \textbf{81.1} & \textbf{84.7} & \textbf{56.0} & \textbf{57.7} \\
\bottomrule
\end{tabular}
\caption{Results of comparative experiments with different methods on multi-hop QA benchmarks. \textbf{Bold} means the best and \underline{underline} means the second. Dagger$^\dagger$ means with GPT-4o.}
\label{tab:main_results}
\end{table}

\begin{table}[!t]
\centering
\small
\setlength{\tabcolsep}{4pt}
\begin{tabular}{l|cc|cc|cc}
\toprule
Dataset &
\multicolumn{2}{c|}{HotpotQA} &
\multicolumn{2}{c|}{2WikiQA} &
\multicolumn{2}{c}{MuSiQue} \\
Metrics &
EM & F1 &
EM & F1 &
EM & F1 \\
\midrule
GPT-3.5-Turbo & 31.9 & 43.7 & 36.0 & 46.6 & 19.2 & 33.3  \\
\midrule
\multicolumn{7}{l}{\cellcolor{gray!15}Module-wise Removal} \\
\midrule
THOR w/o iPFC    & 49.5 & 51.4 & 54.5 & 59.9 & 28.4 & 32.7  \\
THOR w/o iHPC    & 61.2 & 66.2 & 65.7 & 72.3 & 33.1 & 40.2  \\
THOR w/o iACC    & 54.9 & 67.3 & 72.0 & 73.1 & 34.4 & 42.0  \\
THOR w/o memory  & 59.3 & 63.1 & 59.1 & 65.1 & 31.1 & 32.2  \\
\midrule
\textbf{THOR}    & 69.3 & 76.1 & 75.6 & 78.6 & 48.5 & 52.1  \\
\bottomrule
\end{tabular}
\caption{
Ablation study with a GPT-3.5-Turbo backbone on THOR with a removal of individual modules and the slot-schema designed memory.
}
\label{tab:ablation}
\end{table}

\begin{table}[t]
\centering
\small
\setlength{\tabcolsep}{1.5pt}
\renewcommand{\arraystretch}{1}
\begin{tabular}{l|ccc|ccc}
\toprule
\multirow{2}{*}{Method} &
\multicolumn{3}{c|}{FSR(\%) ($\downarrow$)} &
\multicolumn{3}{c}{ASR(\%) ($\downarrow$)} \\
& 2-hop & 3-hop & 4-hop & 2-hop & 3-hop & 4-hop \\
\midrule
CoT & 6.7 & 16.4 & 29.2 & 21.4 & 17.4 & 25.9  \\
CoRAG & 2.1 & 18.2 & 29.4 & 17.0 & 19.7 & 20.7  \\
ReAgent & 2.7 & 12.1 & 21.2 & 19.5 & 12.1 & 21.2  \\
\midrule
THOR w/o iPFC  & 1.6 & 14.9 & 26.1 & 9.8 & 7.7 & 11.6 \\
THOR w/o iHPC  & 1.3 & 12.6 & 18.4 & 16.9 & 14.7 & 14.1 \\
THOR w/o iACC  & 0.9 & 11.5 & 20.7 & 15.7 & 15.4 & 17.8  \\
THOR w/o memory & 1.9 & 12.5 & 22.7 & 25.7 & 25.1 & 31.8  \\
\midrule
\textbf{THOR}           & 0.7 & 9.8 & 15.3 & 7.1 & 8.3 & 10.4 \\
\bottomrule
\end{tabular}
\caption{Validation experiments on THOR with metrics FSR/ASR compared with CoT, CoRAG and ReAgent on MuSiQue dataset.}
\label{tab:val}
\end{table}
\subsubsection{Ablation Study}
Table~2 shows that THOR consistently outperforms the plain GPT-3.5-Turbo backbone on all three benchmarks, indicating that the gains do not come from prompting alone but from the proposed framework. Removing any single component leads to a clear degradation, suggesting that the modules are complementary rather than redundant.
Among the removals, iPFC causes the most severe drop of EM on MuSiQue($\downarrow$20.1); w/o iHPC and w/o iACC also reduce performance substantially across datasets and w/o slot-schema memory has a significant drop($\downarrow$19.9) on the more difficult MuSiQue.

\subsubsection{Attention Decay Mitigation Analysis}
\label{sec:val_ablation_analysis}
We conduct probe-based validation experiments on MuSiQue to verify that THOR effectively mitigates attention decay. Concretely, we compare THOR with three representative major baselines with the best performance among the three categories. To verify that each component mitigates the targeted limitations, we report FSR and ASR on MuSiQue in Table~\ref{tab:val}. Overall, THOR achieves the lowest FSR/ASR across hop depths, with the largest advantages emerging on 4-hop cases, confirming that our design specifically improves robustness as the reasoning chain grows. Comparing module removals, iPFC contributes most directly to reducing frame shift that removing iPFC increases FSR, especially on deeper hops. Unlike one-pass chain-of-thought, iPFC treats the hop plan as a mutable object that can be revised under explicit constraints, enabling controlled backtracking and mitigating error propagation in deep chains. iACC also contributes to lower FSR by detecting inconsistency signals early and triggering repair or replan state instead of letting a wrong trajectory continue. In contrast, removing memory increases FSR as well, suggesting that explicit structured state helps keep the reasoning aligned with the intended decomposition. Anchor shift is most strongly affected by the slot-schema working memory that w/o memory yields a large ASR increase across all hop depths. This indicates that explicit entity binding and canonicalized anchor storage are essential for preventing the reasoning from becoming detached from the intended entity. Meanwhile, iHPC and iACC further reduce ASR by refining retrieval cues and rejecting anchor-missing evidence through verification-driven control, respectively.
\begin{table}[t]
\centering
\small
\setlength{\tabcolsep}{4pt} 
\setlength{\aboverulesep}{2pt}
\setlength{\belowrulesep}{2pt}
\begin{tabular}{l|cc|cc|cc}
\toprule
\multirow{2}{*}{Method} &
\multicolumn{2}{c|}{Reg.} &
\multicolumn{2}{c|}{Adv.} &
\multicolumn{2}{c}{Drop($\downarrow$) (\%)} \\
& 2-hop & 3-hop & 2-hop & 3-hop & 2-hop & 3-hop \\
\midrule
BiDAF   & 43.1 & 46.4 & 34.7 & 32.3 & 19.5 & 30.4 \\
ToR     & 48.2 & 41.5 & 41.4 & 30.9 & 14.2 & 25.6 \\
ReAgent & 68.2 & 62.5 & 59.2 & 50.9 & 13.2 & 18.6 \\
\midrule
\textbf{THOR} & 79.3 & 76.4 & 69.2 & 69.3 & 12.7 & \textbf{9.3} \\
\bottomrule
\end{tabular}
\caption{EM results of additional adversarial document injection experiments on HotpotQA. Drop($\downarrow$) is the relative EM decrease from Reg. to Adv.}
\label{tab:inv_adv}
\end{table}
\subsubsection{Error Accumulation Analysis}
Following the adversarial evaluation protocol that injects documents into the context \citep{jiang-bansal-2019-avoiding}, we construct an adversarial test by appending misleading but topically related documents to each instance to validate the mitigation of error accumulation using EM/F1. The adversarial documents imply that more anchor shifts will occur at various stages of reasoning, resulting in the accumulation of errors and thereby the accuracy drops. As shown in Table~\ref{tab:inv_adv}, we compared THOR with ToR and ReAgent which also enable a repair mechanism. Our method exhibits a much smaller drop (12.7\% and 9.3\%) which indicates that our verification-and-repair control effectively suppresses wrong-path reasoning and subtle anchor missing errors induced by adversarial evidence, supporting our claim that THOR mitigates error accumulation.
\begin{table}[t]
\centering
\small
\renewcommand{\arraystretch}{1.05}
\setlength{\tabcolsep}{7pt}
\begin{tabular}{lccc}
\toprule
Method & HotpotQA & 2Wiki & MuSiQue \\
\midrule
ITER-RETGEN & 50.6 & 51.1 & 27.2 \\
IRCoT       & 46.0 & 46.5 & 25.2 \\
ToR         & 53.1 & 51.8 & 29.5 \\
CoRAG       & 66.0 & 56.5 & 32.2 \\
\midrule
THOR-Iter@1  & 63.1 & 53.0 & 36.4 \\
THOR-Iter@2  & \textbf{73.1} & \textbf{65.8} & 41.5 \\
THOR-Iter@3  & 67.1 & 61.8 & \textbf{44.5} \\
\bottomrule
\end{tabular}
\caption{Results of retrieval accuracy experiments using paragraphs recall@15 on THOR with representative retrieval-augmented methods on multi-hop QA datasets.}
\label{tab:retrieve}
\end{table}
\begin{table*}[t]
\centering
\small
\setlength{\tabcolsep}{3pt}        
\renewcommand{\arraystretch}{0.95} 
\begin{tabular}{lcccccc}
\toprule
\multirow{2}{*}{\begin{tabular}{@{}l@{}}
\textbf{Backbone Model}\\
\scriptsize(Backbones $\rightarrow$ \textbf{THOR})
\end{tabular}}
&
\multicolumn{2}{c}{\textbf{HotpotQA}} &
\multicolumn{2}{c}{\textbf{2Wiki}} &
\multicolumn{2}{c}{\textbf{MuSiQue}} \\
& EM & F1 & EM & F1 & EM & F1 \\
\midrule
\multicolumn{7}{l}{\cellcolor{gray!15}\textbf{Regular Models}} \\
\midrule
GPT-3.5-turbo
  & 31.2 $\rightarrow$ 69.3
  & 46.9 $\rightarrow$ 76.1
  & 43.2 $\rightarrow$ 75.6
  & 46.1 $\rightarrow$ 78.6
  & 19.7 $\rightarrow$ 48.5
  & 24.5 $\rightarrow$ 52.1 \\
Llama-4-Instruct
  & 26.3 $\rightarrow$ 53.1
  & 38.9 $\rightarrow$ 58.3
  & 33.2 $\rightarrow$ 59.1
  & 46.7 $\rightarrow$ 62.3
  & 10.7 $\rightarrow$ 31.1
  & 18.5 $\rightarrow$ 39.9 \\
DeepSeek-V3
  & 35.2 $\rightarrow$ 74.7
  & 49.1 $\rightarrow$ 76.1
  & 46.6 $\rightarrow$ 75.1
  & 57.9 $\rightarrow$ 81.4
  & 22.3 $\rightarrow$ 49.1
  & 33.0 $\rightarrow$ 57.2 \\
Qwen-2.5-Instruct
  & 36.3 $\rightarrow$ 63.7
  & 51.9 $\rightarrow$ 74.5
  & 54.3 $\rightarrow$ 65.6
  & 63.1 $\rightarrow$ 78.3
  & 22.2 $\rightarrow$ 34.2
  & 32.7 $\rightarrow$ 43.6 \\
Gemini-1.5-Flash
  & 37.4 $\rightarrow$ 59.1
  & 48.8 $\rightarrow$ 68.2
  & 56.3 $\rightarrow$ 67.2
  & 65.0 $\rightarrow$ 72.3
  & 20.8 $\rightarrow$ 30.8
  & 31.0 $\rightarrow$ 42.1 \\
Gemini-2.0-Flash
  & 37.1 $\rightarrow$ 62.1
  & 49.0 $\rightarrow$ 68.4
  & 53.8 $\rightarrow$ 75.4
  & 65.1 $\rightarrow$ 77.9
  & 24.6 $\rightarrow$ 29.1
  & 33.8 $\rightarrow$ 41.6 \\
GPT-4o
  & 38.1 $\rightarrow$ 72.1
  & 54.9 $\rightarrow$ \textbf{81.4}
  & 51.7 $\rightarrow$ 81.1
  & 64.9 $\rightarrow$ 84.7
  & 24.5 $\rightarrow$ 56.0
  & 37.9 $\rightarrow$ 57.7 \\
GPT-4.1
  & 38.9 $\rightarrow$ \textbf{73.2}
  & 56.3 $\rightarrow$ 79.2
  & 54.4 $\rightarrow$ 78.4
  & 66.5 $\rightarrow$ 85.9
  & 27.1 $\rightarrow$ 49.6
  & 41.3 $\rightarrow$ 59.3 \\
\midrule
\multicolumn{7}{l}{\cellcolor{gray!15}\textbf{Reasoning Models}} \\
\midrule
DeepSeek-R1
  & 35.6 $\rightarrow$ 61.2
  & 48.3 $\rightarrow$ 73.8
  & 60.1 $\rightarrow$ 62.3
  & 70.7 $\rightarrow$ 69.2
  & 29.8 $\rightarrow$ 34.5
  & 41.6 $\rightarrow$ 45.7 \\
Qwen-3-Thinking
  & 36.1 $\rightarrow$ 64.1
  & 50.6 $\rightarrow$ 69.2
  & 62.4 $\rightarrow$ 67.4
  & 72.9 $\rightarrow$ 73.0
  & 27.1 $\rightarrow$ 41.5
  & 38.7 $\rightarrow$ 46.7 \\
Gemini-2.5-Pro
  & 43.0 $\rightarrow$ 65.6
  & 56.0 $\rightarrow$ 58.3
  & 74.3 $\rightarrow$ 80.1
  & 82.9 $\rightarrow$ 82.3
  & 38.3 $\rightarrow$ 45.1
  & 49.1 $\rightarrow$ 59.8 \\
O1
  & 50.5 $\rightarrow$ 53.4
  & 66.1 $\rightarrow$ 69.2
  & 65.6 $\rightarrow$ 76.2
  & 75.8 $\rightarrow$ 81.5
  & 41.7 $\rightarrow$ 39.1
  & 55.1 $\rightarrow$ 49.4 \\
O3
  & 53.5 $\rightarrow$ 69.0
  & 69.6 $\rightarrow$ 78.2
  & 70.6 $\rightarrow$ \textbf{85.6}
  & 78.7 $\rightarrow$ \textbf{90.4}
  & 44.2 $\rightarrow$ \textbf{56.1}
  & 57.9 $\rightarrow$ \textbf{63.2} \\
\bottomrule
\end{tabular}
\caption{Backbone adaptation experiment results. Each cell reports baseline $\rightarrow$ THOR performance for EM and F1 on three multi-hop QA datasets. \textbf{Bold} indicates the best result in all the experiments}
\label{tab:backbone-adaptation}
\end{table*}
\subsubsection{Retrieval Accuracy Analysis}
\label{sec:ablation_retrieval}
To isolate whether THOR improve better evidence retrieval, we further evaluate retrieval quality using recall@15 in ~\citep{trivedi-etal-2023-interleaving}. In Table~\ref{tab:retrieve}, we compared representative retrieval-augmented methods with THOR while THOR-Iter@$k$ denotes allowing at most $k$ rounds of hop-local retrieval refinement within our topic-aware + context-aware retriever.
Overall, THOR iterative refinement substantially improves retrieval recall compared to prior multi-hop retrievers.
On the most challenging dataset MuSiQue, recall increases from 36.4 (Iter@1) to 44.5 (Iter@3), yielding a 12.3\% gain over CoRAG and this indicates that deeper reasoning particularly need iterative, failure-aware refinement which also proves that THOR benefits from additional refinement rounds.
\subsubsection{Adaptation Analysis}
\label{sec:adaptation}
A key claim of THOR is generalization that it should function as a plug-and-play reasoning wrapper that can adapt to different LLM backbones. To test this, we keep the THOR fixed and only swap the backbone model used to instantiate the modules. Table~\ref{tab:backbone-adaptation} reports performance from single model to THOR with same backbones.
Across a wide range of regular LLMs, THOR consistently yields large gains on all three datasets, often transforming weak baselines into strong multi-hop solvers. These results indicate that THOR maximizes the superior capabilities of LLMs that its benefits arise from the integrative combination of the framework-level control and intrinsic ability. In particular, for more complex dataset MuSiQue, THOR results in better performance with reasoning models, especially with o3 model.
\subsubsection{Accuracy-Cost Trade-off Analysis}
\begin{figure}
    \centering
    \includegraphics[width=1\linewidth]{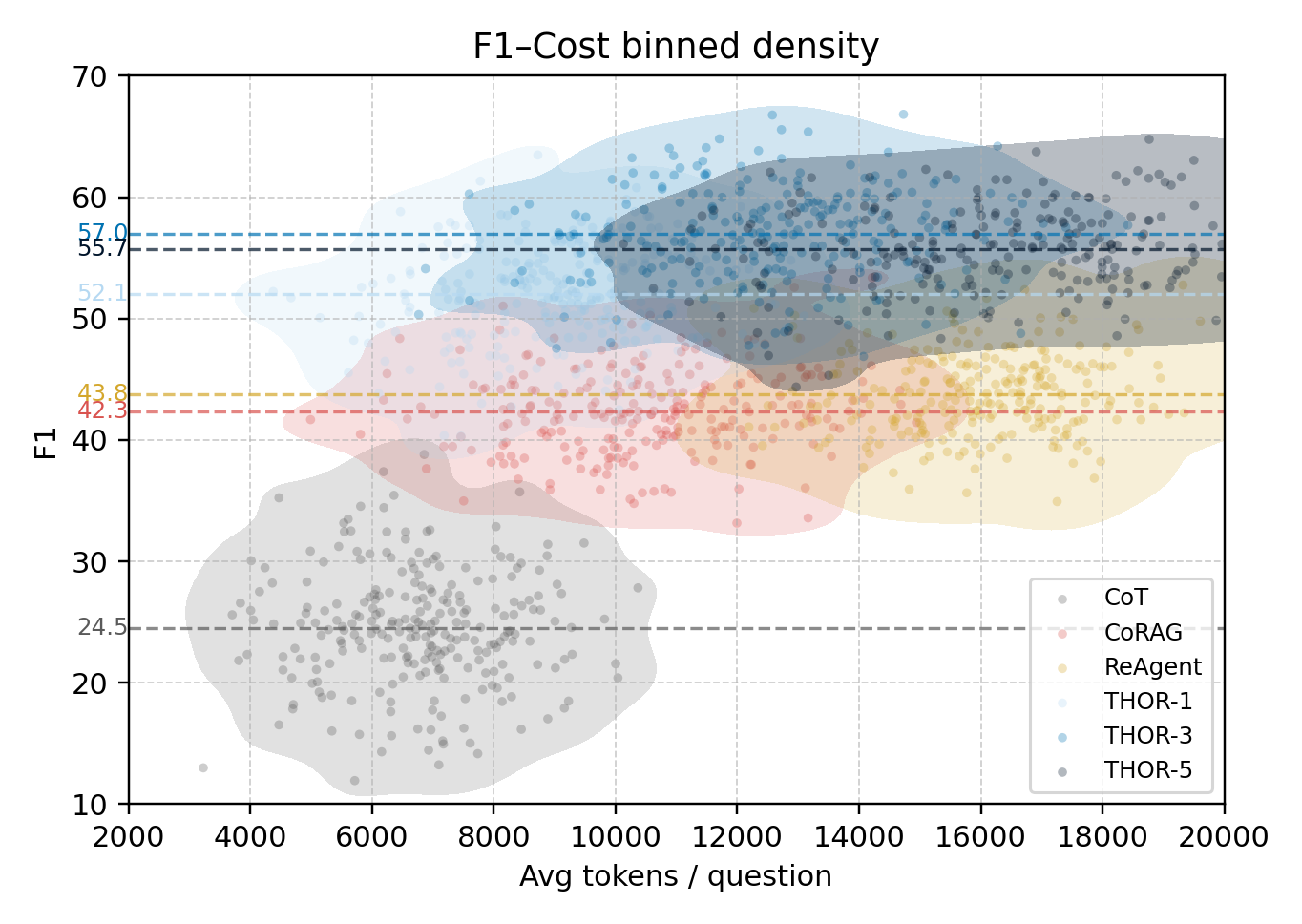}
    \caption{Accuracy-cost trade-off experiments on THOR with CoT, CoRAG and ReAgent on MuSiQue dataset using F1 metrics.}
    \label{fig:acccost}
\end{figure}

We used 10-binned density distribution to illustrate the relationship between accuracy and cost. To make this trade-off explicit, we sweep THOR per-hop retry budget $I_{\max}=1,3,5$ on MuSiQue. We measure cost as the average total tokens per question, and compare against CoT, CoRAG, ReAgent under the same retrieval resources. As shown in Fig.~\ref{fig:acccost}, CoT concentrates in a low-cost/low-accuracy region, while CoRAG trades additional tokens for moderate gains. Compared to ReAgent with multi-agent systems settings, THOR-3 performs better F1 with fewer tokens. THOR forms a controllable accuracy--cost frontier that increasing $I_{\max}$ consistently shifts the operating point rightward (higher token budget) and upward (higher EM/F1), reflecting additional targeted retrieval refinement and local/global corrections.

\label{sec:acc_cost}

\section{Discussion}

\subsection{Controlled add-on experiment: same modules, different control}
To identify where THOR's improvement comes from, we conduct a supplementary experiment as shown in ~\ref{tab:addon} in which we keep the same planner (iPFC) and the same retriever (iHPC) across conditions, and vary only (a) the control protocol and (b) the global-state representation on MuSiQue.

\begin{table*}[t]
\centering
\small
\resizebox{\linewidth}{!}{
\begin{tabular}{l l l c c c c}
\toprule
\textbf{Condition} & \textbf{Control Protocol} & \textbf{Global State} & \textbf{EM$\uparrow$} & \textbf{F1$\uparrow$} & \textbf{FSR$\downarrow$} & \textbf{ASR$\downarrow$} \\
\midrule
Standard planner--executor loop & Plan $\rightarrow$ Execute & Text scratchpad & 31.3 & 39.9 & 17.1 & 20.2 \\
+ Typed Frame only & Plan $\rightarrow$ Execute & Slot-schema frame & 43.3 & 49.6 & 15.6 & 10.8 \\
+ Controller only & $\theta \leftrightarrow \gamma$ finite-state control & Text scratchpad & 39.2 & 46.0 & 13.1 & 15.7 \\
THOR (full) & $\theta \leftrightarrow \gamma$ finite-state control & Slot-schema frame & \textbf{48.5} & \textbf{52.1} & \textbf{12.1} & \textbf{8.8} \\
\bottomrule
\end{tabular}
}
\caption{THOR (full) outperforms the standard control setting on EM/F1 while reducing FSR/ASR. This indicates that the benefit comes from the combination of the control protocol, the typed frame, and verification-as-control, rather than from simple module stacking. The ``Typed Frame only'' condition isolates the contribution of constraint binding for reducing local anchor shift, while the ``Controller only'' condition isolates the contribution of explicit backtracking for reducing global frame shift.}
\label{tab:addon}
\end{table*}

\subsection{Why THOR works at a deeper level}

\paragraph{Key problem insight.}
Multi-hop QA is a sequential decision process that requires two coupled functions to remain reliable over long reasoning chains:
(i) maintaining a stable task representation (the core entity, constraints, and expected answer type), and
(ii) executing local evidence reasoning at each hop.
As the context grows, the model becomes prone to two observable failure patterns:
Attention decay: intermediate sub-questions gradually drift away from the original intent, manifesting as frame shift and anchor shift. In practice, this appears as locally plausible hops that are no longer consistent with the global constraints.
Error accumulation: early misbindings or unsupported intermediate conclusions propagate forward. Later hops may remain internally coherent while being conditioned on a wrong premise, making the final answer confidently wrong and difficult to recover with generic ``reflect-and-retry.''
\paragraph{How THOR resolves this by design.}
THOR enforces a two-timescale protocol in which the $\theta$ phase serves as an outer-loop controller that periodically re-stabilizes the task representation by reasserting the core entity, type, and constraints, repairing misalignment, and deciding whether global replanning/backtracking is needed. The $\gamma$ phase serves as an inner-loop executor specialized for local hop work under the stabilized frame. Crucially, THOR operationalizes mismatch-triggered corrective control: structured verification signals drive explicit state transitions that escalate from local refinement to global repair, replan, or backtracking. In this way, THOR closes the loop between prediction and evidence and prevents drift and compounding errors from silently accumulating.

THOR performs robustly in the processes of breaking down and retrieving information on various fine-grained issues, while handling and interpreting the details of the information can be challenging.

\section{Conclusion}
\label{sec:conclusion}
We presented \textbf{THOR}, a brain-inspired Theta--Gamma hierarchical oscillatory reasoning framework for multi-hop question answering. THOR addresses failures by decoupling global planning from local retrieval, enabling efficient attention transfer between hops and a verification and repair mechanism that interrupts the accumulation of errors in the wrong paths. Comprehensive experiments demonstrate higher accuracy, robustness and generalization of our method. Future work focuses on delving deeper into human brain reasoning and further improving the framework.

\section*{Limitations}
We observe failures where the provided supporting evidence contains only minor lexical or contextual differences, making the inconsistency difficult to detect and leading the model to commit to an incorrect hop. Simultaneously, errors also arise from imperfect evidence selection that insufficient filtering allows weakly related, ultimately confuses the model during multi-hop aggregation.
\section*{AI Assistance}
We used AI assistants to support language polishing and minor code debugging. All technical content and conclusions were verified by the authors.
\section*{Ethics}
This work uses only publicly available benchmark datasets for multi-hop question answering, all released for research purposes. We use these artifacts under their respective licenses/terms.
\section*{Acknowledgments}
This research was supported by Gusu Innovation and Entrepreneurship Leading Talent Program Project (Grant No.ZXL2024349).
\bibliography{custom}

\appendix
\section{Analysis of Previous Works}
\subsection{Limitations of Previous Methods}
\label{sec:app_error_analysis}
\begin{figure}[t]
    \centering
    \includegraphics[width=1\linewidth]{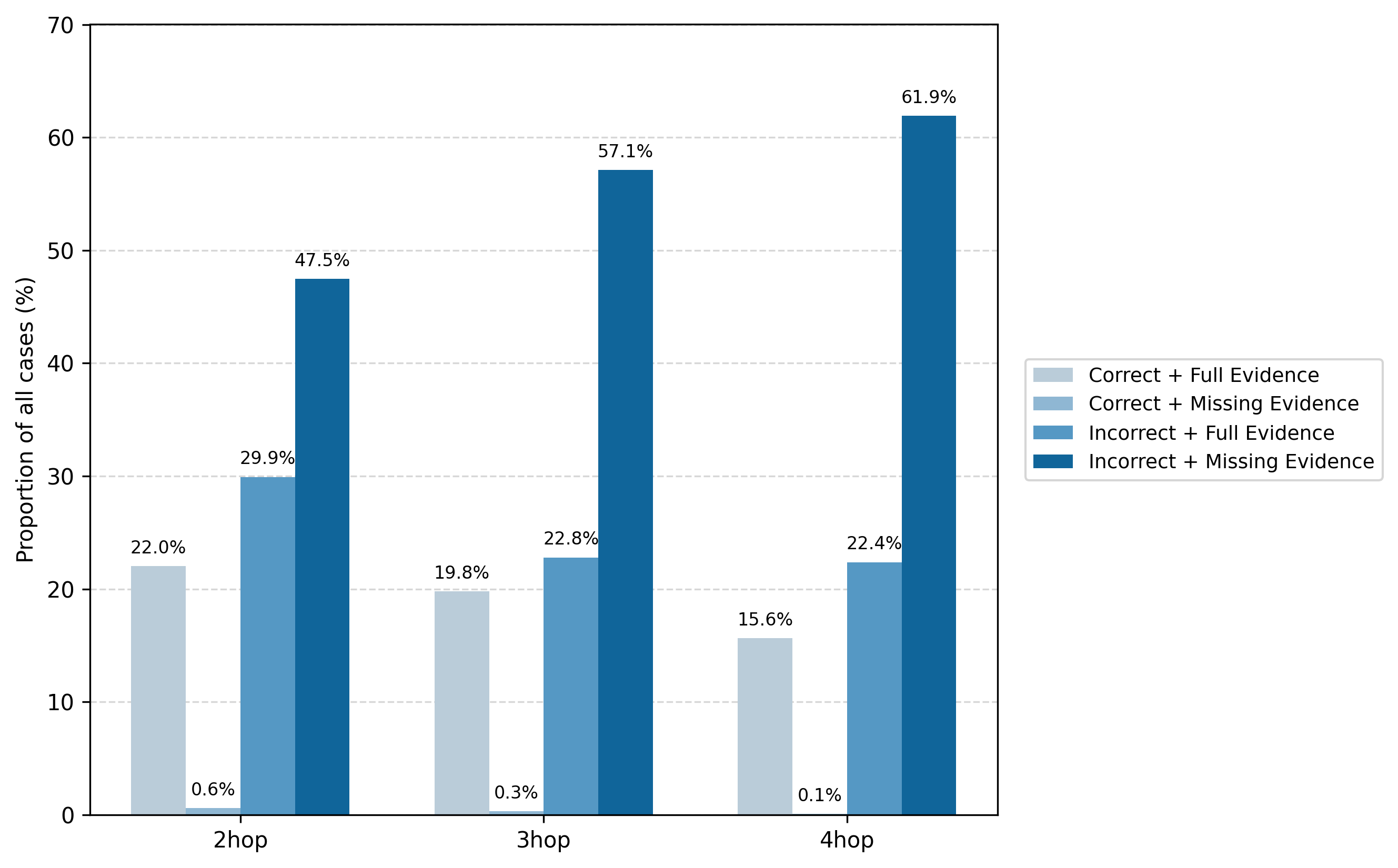}
    \caption{The proportion of \textit{incorrect but with all evidence present} samples remains relatively high on MuSiQue, revealing a faithfulness gap between evidence availability and conclusion integration.}
    \label{fig:iv}
\end{figure}

As shown in Fig.~\ref{fig:iv}, we use Chain-of-Thought~\citep{wei2023chainofthoughtpromptingelicitsreasoning} strategy to implementation comprehensive test on MuSiQue dataset. Prior multi-hop QA systems can be broadly grouped into three categories—prompt-engineering pipelines, retrieval-augmented reasoning, and agent-/multi-agent-based frameworks. Although each category offers partial remedies, none provides a stable, explicit, and globally consistent mechanism to (i) detect wrong-path reasoning, (ii) isolate and suppress error accumulation across hops, and (iii) repair or replan under verifiable constraints. We summarize the key limitations below.
\paragraph{Prompt-engineering pipelines: shallow control and strong dependence on the backbone.}
Prompt-centric improvements (e.g., chain-of-thought style decomposition, self-consistency, reflection prompts, or hand-crafted templates) are typically \emph{lightweight wrappers} over the backbone model.
As a result, their effectiveness is tightly coupled to the inherent reasoning and instruction-following capacity of the underlying LLM: stronger models benefit more, while weaker/cheaper models often exhibit unstable gains.
Moreover, prompt pipelines rarely expose a structured and auditable state (e.g., an explicit frame memory with immutable validated prefix); they rely on implicit attention allocation inside the model.
This makes them vulnerable to long-context interference: as the hop chain grows, earlier constraints and intermediate commitments become less salient, leading to progressive frame drift and type confusion.
In practice, prompt tuning can improve surface-level coherence, but does not provide a principled way to separate search errors (missing evidence) from reasoning errors (mis-integration given correct evidence), nor does it offer deterministic control transitions for repair vs.\ replan.
\paragraph{Retrieval-augmented reasoning: better evidence, but not necessarily better answers.}
RAG-style methods can substantially improve retrieval quality, yet multi-hop QA failures persist even when the system retrieves the correct supporting documents.
A common phenomenon is the faithfulness gap: the model may have access to all necessary evidence but still produce an incorrect final answer due to erroneous integration, distraction by plausible but irrelevant facts, or over-reliance on priors. Furthermore, retrieval does not guarantee entity anchoring and logical consistency across hops.
Even with correct evidence in the candidate pool, the model may silently shift the anchor entity (entity drift), mismatch the expected answer type, or compose hop results into an inconsistent chain. These errors are especially frequent when multiple entities share similar surface forms or when intermediate answers are underspecified.
Critically, most retrieval-augmented systems lack an explicit verifier that can attribute failure to (a) evidence insufficiency, (b) incorrect hop framing, or (c) global-plan inconsistency; consequently, their fallback strategy is often a generic retrieve more loop, which increases cost without reliably resolving reasoning failures.
\paragraph{Agent-/multi-agent-based methods: emerging capability, but unstable convergence and limited global reflection.}
Agent-based frameworks introduce modularity (planner, retriever, verifier, etc.) and can repair certain local errors, but typical multi-agent systems still face three structural weaknesses.
First, control is often implicit: interactions are mediated through natural language messages without a formally defined state machine or typed verification signals, which makes convergence behavior sensitive to stochasticity, prompt phrasing, and model variance.
Second, repairs are frequently myopic: agents can revise the current hop output, but they often lack a globally consistent frame object that is explicitly maintained, audited, and minimally updated under constraints (i.e., no principled global frame repair and replan).
Third, many agentic systems do not implement retrieval-failure path replay: when retrieval fails, they re-query heuristically rather than replaying the failure trajectory to diagnose whether the issue comes from a wrong anchor, an incorrect sub-question, or a contradiction created earlier in the chain.
As a result, multi-agent loops may oscillate, over-consume tokens, or repair locally while the overall chain remains logically inconsistent.

As shown in table~\ref{tab:pe} and ~\ref{tab:ma}, THOR differs by detecting error and correcting it, turning multi-hop QA into a closed-loop control system with explicit $\theta \leftrightarrow \gamma$ scheduling, typed global constraints, and diagnosis-driven repair. This makes behavior more reproducible and correctable, rather than relying on heuristic prompting.

\begin{table*}[t]
\centering
\small
\begin{tabularx}{\textwidth}{lXX}
\toprule
\textbf{Dimension} & \textbf{Planner--Executor (typical)} & \textbf{THOR (ours)} \\
\midrule
Control policy & Implicit prompt-driven loop & Explicit two-timescale finite-state controller \\
Global constraints & Best-effort via long context; prone to error accumulation & Enforced via slot-schema frame alignment/repair at every hop \\
Retry/repair & Generic reflect & Diagnosis-driven via a structured check vector \\
Auditability & Unstructured traces; hard to pinpoint where drift begins & Loggable state transitions + slot-schema memory: what shifted and why \\
\bottomrule
\end{tabularx}
\caption{Comparison between THOR and Planner--Executor.}
\label{tab:pe}
\end{table*}

\begin{table*}[t]
\centering
\small
\begin{tabularx}{\textwidth}{lXX}
\toprule
\textbf{Dimension} & \textbf{Modular Agents} & \textbf{THOR (ours)} \\
\midrule
Orchestration & Modules wired with ad hoc glue logic & Unifying $\theta \leftrightarrow \gamma$ control protocol with explicit controller actions and transition rules \\
Memory / global state & Often textual or untyped memory/scratchpad & Typed slot-schema frame encoding entities and constraints \\
Verifier role & Often a score/critique used post hoc & Verification-as-control: structured diagnostics directly gate state transitions \\
Correction behavior & Often ``try again'' or rewrite steps; escalation is heuristic & Targeted backtracking/repair/replan guided by diagnostic checks \\
\bottomrule
\end{tabularx}
\caption{Comparison between THOR and Modular Agents.}
\label{tab:ma}
\end{table*}

Figure~\ref{fig:iv} shows that the proportion of incorrect but with all evidence present cases remains non-trivial on MuSiQue, indicating that improving retrieval recall alone does not guarantee faithful answer synthesis.
\label{sec:app_prev_methods_limits}
This appendix provides additional analyses and implementation details that support the main paper. We focus on two failure families, attention decay and error accumulation.
\subsection{Attention Decay}
\label{sec:app_error_examples}
\begin{figure}[t]
    \centering
    \includegraphics[width=1\linewidth]{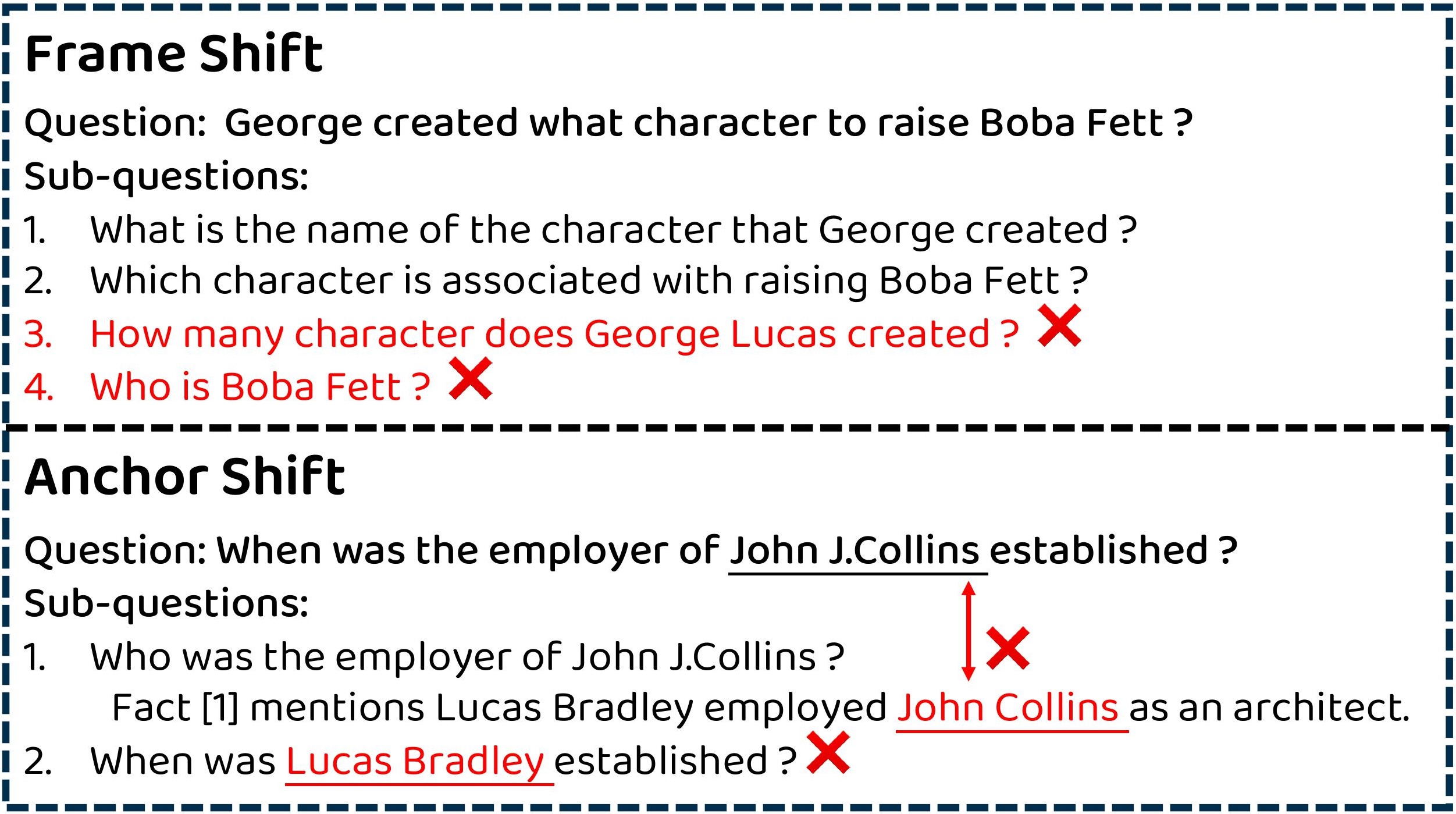}
    \caption{Examples of two representations of attention decay: (left) \textbf{Frame Shift} (global decomposition drift) and (right) \textbf{Anchor Shift} (entity-level drift under a seemingly correct frame).}
    \label{fig:ad}
\end{figure}

Figure~\ref{fig:ad} illustrates two common attention-decay patterns in multi-hop reasoning: (i) Frame Shift, where the reasoning trajectory deviates from the intended decomposition plan and enters a logically disoriented chain; and (ii) Anchor Shift, where the high-level frame remains plausible but the retrieved evidence becomes grounded on a wrong or overly similar entity. We expand both patterns below with concrete manifestations and typical causes.
\paragraph{Frame Shift}
Frame Shift refers to a progressive divergence between the original question objective and the current hop objective. It often appears as one or more of the following observable behaviors:
(1) Hop objective drift: the generated sub-question begins to optimize for a different goal than what the remaining chain requires, even though the sub-question is fluent and answerable in isolation.
(2) Constraint drop: type, temporal, or relational constraints implied by earlier hops disappear in later prompts or later retrieval cues, so the chain loses the intended direction.
(3) Inconsistent intermediate commitments: the chain commits an intermediate answer that is compatible with the local hop but incompatible with the global plan, and subsequent hops treat this commitment as a premise rather than checking plan consistency.
(4) Spurious bridge creation: the system selects an intermediate entity that creates an easy but incorrect bridge to later hops, producing a chain that looks coherent yet does not support the final answer.
(5) Plan discontinuity: the system unexpectedly changes the decomposition structure, such as merging two hops into one, skipping a necessary hop, or introducing an unrelated hop that cannot be mapped back to the original plan.
Frame Shift is typically caused by a combination of long-context interference and weak plan invariants:
(1) Prefix fading under long context: as the chain grows, earlier plan tokens, intermediate answers, and constraints become less salient relative to newly retrieved passages, causing the model to re-interpret what the next hop should achieve.
(2) Local optimality bias: the model prefers a locally answerable hop question that yields a confident sub-answer, even if that hop does not advance the global objective.
(3) Error propagation from early decomposition: an early subtle decomposition error is treated as correct, and later hops inherit the wrong premise; because the later hops are conditioned on the wrong premise, they become self-reinforcing.
(4) Ambiguous bridging entities: when multiple plausible intermediate entities exist, the model may select a convenient bridge that shortens reasoning but breaks the required logical chain.
(5) Lack of explicit plan validation: without a controller that enforces immutable validated prefix and checks plan consistency at each hop, the system has no reliable mechanism to detect and correct plan drift.
\begin{figure}[t]
    \centering
    \includegraphics[width=1\linewidth]{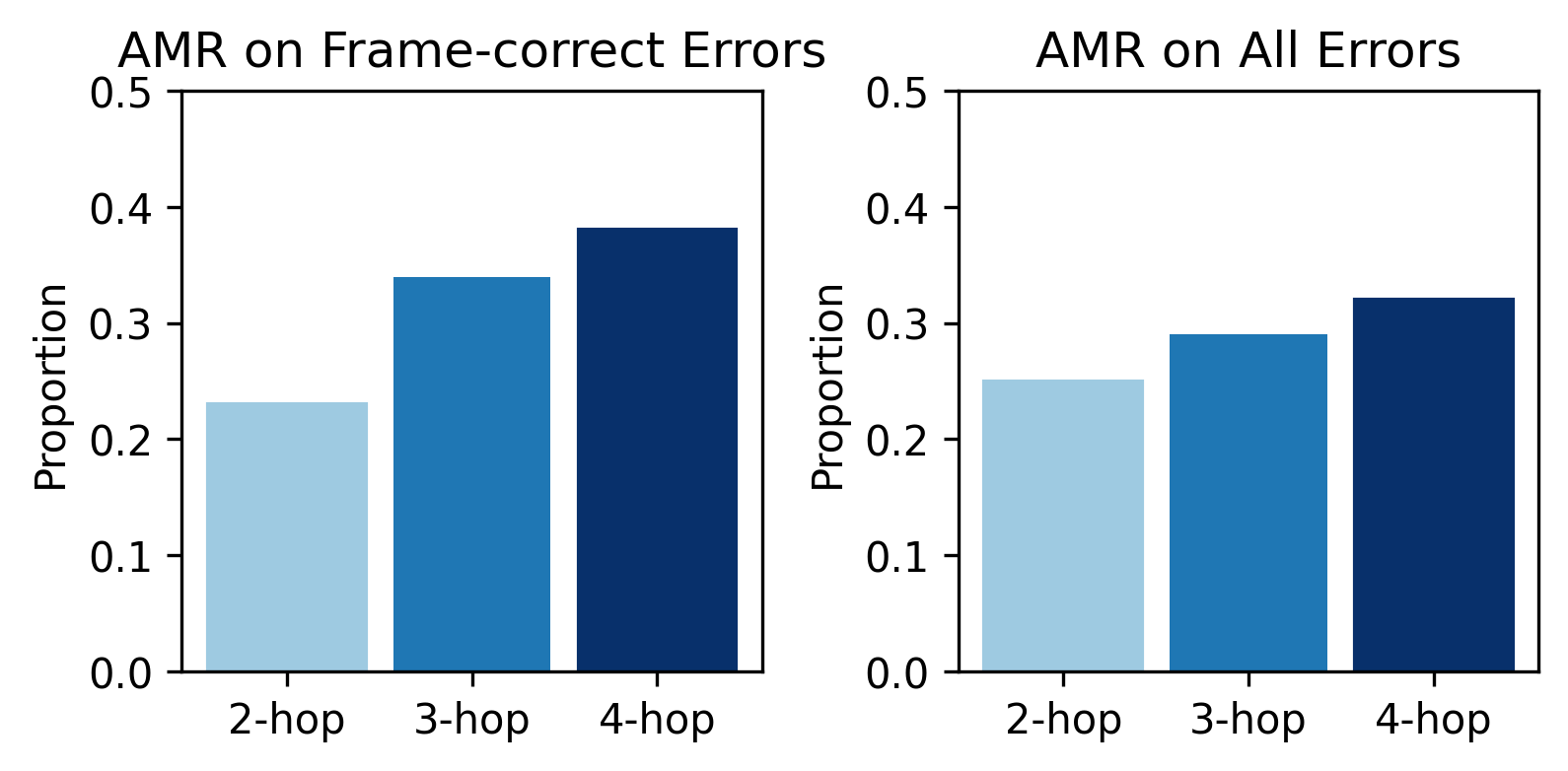}
    \caption{\textbf{ASR} across hop settings on frame-correct but answer-error cases and all error cases.}
    \label{fig:amr}
\end{figure}
\paragraph{Anchor Shift}
Anchor Shift refers to a progressive deviation of the entity grounding for retrieval and integration while the chain-level goal remains superficially aligned. As shown in Fig.~\ref{fig:amr}, it appears all position on MuSiQue. It often appears as:
(1) Entity substitution: the chain replaces the intended core entity with a near-duplicate, homonym, acronym variant, or similarly named entity.
(2) Alias overgeneralization: retrieval cues expand to an alias that matches multiple entities, and the system implicitly commits to the wrong referent.
(3) Context-induced hijacking: a retrieved passage contains a prominent related entity that attracts subsequent retrieval and reasoning, gradually replacing the original anchor.
(4) Title-level mismatch: the evidence titles appear relevant to the current sub-question, but the underlying article is about a different entity with an overlapping surface form.
(5) Stable sub-questions with unstable grounding: sub-questions remain consistent with the plan textually, yet the actual evidence and intermediate answers are grounded to a different entity than intended. Anchor Shift is strongly tied to retrieval conditioning and entity ambiguity:
(1) Surface-form ambiguity: Wikipedia titles and entity mentions often share identical
\subsection{Error Accumulation}
\label{sec:app_faithfulness_gap}
\paragraph{Definition.}
We define error accumulation in multi-hop reasoning as the phenomenon where small inaccuracies introduced at earlier hops progressively propagate, amplify, and entangle with later decisions, such that the final prediction becomes unreliable even if each hop appears locally plausible. Error accumulation refers to the increasing probability that later hops become incorrect conditioned on earlier deviations, together with the observation that the magnitude of downstream degradation increases as the error is repeatedly committed into memory, yielding a compounding effect rather than an isolated local mistake.
\begin{figure*}[t]
\centering
\begin{tcolorbox}[
  width=\textwidth,
  colback=gray!6,
  colframe=black,
  boxrule=0.4pt,
  arc=2pt,
  left=6pt,right=6pt,top=6pt,bottom=6pt
]
\small
\textbf{Example (Error Accumulation Across Hops).}
Consider the question:
\emph{``Which university did the author of \textit{The Hobbit} attend?''}

\textbf{Hop 1 (anchor selection error).}
The system decomposes into $sq_1$:
\emph{``Who is the author of \textit{The Hobbit}?''}
It retrieves a page about \textbf{Peter Jackson} (director of \textit{The Lord of the Rings} films) and incorrectly commits:
\[
a_1 = \text{``Peter Jackson''}, \quad ce_1=\text{Peter Jackson}.
\]
This looks locally plausible because the evidence mentions \textit{The Hobbit} (film) and prominent names.

\textbf{Hop 2 (query drift caused by the committed premise).}
Using the committed $a_1$, it forms $sq_2$:
\emph{``Which university did Peter Jackson attend?''}
Retrieval now focuses on Peter Jackson biographies and returns evidence about his education.
The system answers:
\[
a_2 = \text{``(some institution)''}.
\]

\textbf{Hop 3 (locally coherent but globally wrong chain).}
A verifier that only checks local grounding may accept Hop 2 because the evidence does support Peter Jackson's education.
However, the chain is already irrecoverably off-target: the original question asks about \textbf{J.\,R.\,R. Tolkien}, not Peter Jackson.

\textbf{Why this is \emph{error accumulation}.}
The initial mistake at Hop 1 is small (confusing book author with film-related entity), but it is \emph{written into the chain} and reused as a premise for later hops.
Downstream hops become increasingly consistent with the wrong anchor, making the trajectory:
\[
\epsilon_1 \Rightarrow \epsilon_2 \Rightarrow \epsilon_3
\]
both \emph{self-reinforcing} and \emph{hard to detect} from any single hop.
\end{tcolorbox}
\caption{Illustration of error accumulation across hops. A locally plausible anchor error at the first hop is committed into the reasoning chain and propagates through later sub-questions, yielding a globally incorrect but locally coherent trajectory.}
\label{fig:error_accumulation_box}
\end{figure*}
\paragraph{Mechanism: why accumulation happens.}
Error accumulation arises from two coupled mechanisms: (1)Stateful dependence across hops. Later decisions explicitly depend on earlier outputs through plan slots, intermediate answers, and retrieval queries (e.g., $ce_{t+1}$ and $sq_{t+1}$ are functions of $o_t$). Once an incorrect output is used as input, the system optimizes its search around an incorrect region of the hypothesis space; (2)Long-context interference. As the reasoning chain grows, earlier information competes with newly retrieved contexts. This increases the chance that the model implicitly reinterprets earlier commitments (``soft forgetting''), creating inconsistent internal states that further destabilize later hops.
\paragraph{Observable signatures.}
In practice, error accumulation can be diagnosed by:
(i) monotonic degradation of hop-wise verification scores as $t$ increases,
(ii) increasing inconsistency between expected type $\tau_t$ and produced sub-answers,
(iii) divergence between retrieved evidence sets and the intended chain goal (frame drift),
and (iv) failure cases where all gold evidence is present but the final answer remains incorrect (faithfulness gap), indicating reasoning corruption rather than retrieval failure.

\section{Human Brain Inspiration}
\label{sec:app_ins}
\subsection{Human Brain Mechanisms}
\begin{figure}
    \centering
    \includegraphics[width=1\linewidth]{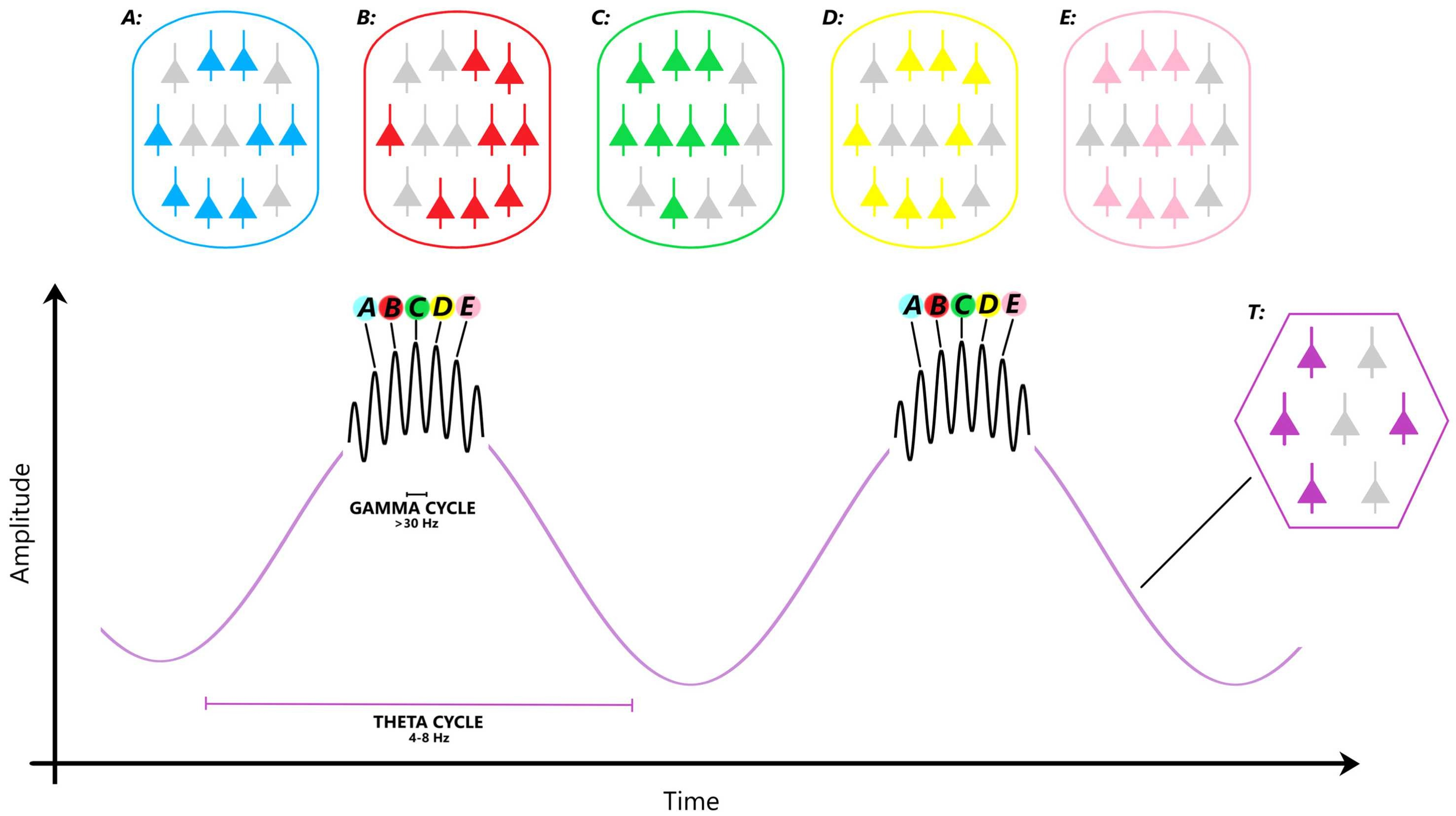}
    \caption{Theta--Gamma hierarchical oscillation in human brain.}
    \label{fig:apptg}
\end{figure}

As shown in Figure ~\ref{fig:apptg}, Theta (approximately 3--8 Hz in humans, with partially overlapping bands across species and tasks) and Gamma (approximately 30--140 Hz, often subdivided into low/high gamma) are prominent neuronal oscillations observed in hippocampal--cortical circuits. A widely supported organizational principle is cross-frequency coupling, in which the phase of theta modulates the amplitude or timing of gamma activity, commonly referred to as theta--gamma phase--amplitude coupling (PAC). In electrophysiological recordings, gamma bursts tend to occur at specific phases of the theta cycle, yielding a temporal structure that organizes neuronal spiking and local synaptic integration within repeated theta cycles.
A key functional role of theta rhythm is large-scale coordination across distributed brain regions. Theta oscillations can exhibit inter-regional phase synchrony or phase locking, providing a shared temporal reference that aligns the excitability windows of neuronal populations across areas. Such alignment is frequently reported between the hippocampus and prefrontal cortex during memory-guided behavior, decision-making, and cognitive control, and it is also observed between hippocampal subfields and connected cortical regions. Within this coordinated regime, theta phase can gate when information is preferentially encoded, retrieved, or transmitted, while gamma activity reflects more local computations, including feature binding, assembly formation, and high-resolution representation within a region.
Within the hippocampus, theta rhythm is strongly influenced by the medial septum and related subcortical inputs, which modulate hippocampal interneuron networks and help pace rhythmic inhibition/excitation. This pacing shapes windows of enhanced neuronal excitability, thereby constraining when principal cells are most likely to fire. Nested within these theta-defined windows, gamma bursts are associated with transient synchronization of local neuronal ensembles and can reflect distinct input streams; for example, gamma-band activity in hippocampal circuits is linked to interactions among CA1, CA3, and entorhinal cortex, with different gamma sub-bands frequently associated with different pathways and computational modes. Although the mapping between specific gamma sub-bands and pathways depends on experimental context, the general observation is that gamma synchronization supports local processing and selective routing of information~\citep{FiebelkornKastner2019,LismanJensen2013,FisahnEtAl1998,Buzsaki2002}.
Cross-regional regulation is further modulated by neuromodulatory systems. Cholinergic, noradrenergic, and dopaminergic inputs can reshape oscillatory dynamics by altering neuronal excitability, synaptic gain, and the balance of inhibition and excitation. In particular, cholinergic signaling is often associated with enhancing rhythmic coordination and stabilizing network states conducive to encoding and attentional engagement, while other neuromodulators contribute to state-dependent reconfiguration of coupling strength and communication efficacy. Consequently, theta--gamma coupling is not a fixed property but a dynamic control mechanism whose strength, preferred phase relationship, and spatial extent vary with task demands and behavioral state~\citep{LismanJensen2013,JonesWilson2005,SchultzDayanMontague1997,FosterWilson2007}.
Overall, theta oscillations provide a temporally structured backbone for inter-areal coordination, while gamma activity supports fast local computations that are organized within theta cycles. Theta--gamma coupling offers a mechanistic substrate for multiplexing information across time, gating communication between regions, and coordinating distributed neural assemblies under changing cognitive states.
\subsection{Mechanisms for Sustaining Attention}
Theta--Gamma Hierarchical Oscillation prevents attention attenuation through three synergistic dimensions:
(1) \textit{Rhythmic Resetting}: Theta oscillations act as a ``master clock,'' providing periodic reset signals that refresh attentional resources at the end of each reasoning hop, preventing resource exhaustion;
(2) \textit{Hierarchical Priority Allocation}: Gamma amplitude varies dynamically across the theta phase. The rising edge facilitates high-intensity encoding, the peak region filters redundant features, and the falling edge focuses on result integration;
(3) \textit{Dynamic Resource Scheduling}: Mediated by the cholinergic system, the brain achieves spatial scheduling. High acetylcholine levels stabilize gamma for active computation, while the Medial Septum (MS) modulates hippocampal theta to facilitate cross-step information transmission without overloading a single region.
\subsection{Strategies for Suppressing Error Accumulation.}
To ensure reasoning accuracy across multiple steps, THOR implements a four-fold suppression strategy:
(1) \textit{Step Isolation}: Multiple gamma subcycles nested within a single theta cycle encode separate reasoning steps. Temporal separation and lateral inhibition between these subcycles prevent cross-interference and error propagation;
(2) \textit{Predictive Monitoring}: The Prefrontal Cortex (PFC) sends predictive signals via theta rhythm to be compared with real-time hippocampal gamma encoding. Mismatches trigger theta phase shifts and dopamine release to initiate immediate correction;
(3) \textit{Closed-Loop Feedback}: A ``computation--verification--adjustment'' circuit between the PFC, hippocampus, and MS operates within each theta cycle, ensuring errors are corrected before they enter subsequent reasoning stages;
(4) \textit{Structured Storage}: Information is stored as ``theta phase-locked gamma sequences.'' These ordered memory traces provide a structured basis for the brain to systematically retrace and correct accumulated errors after the reasoning process is complete.
\section{Experiments Details}
\label{sec:app_exp}
\subsection{Dataset Details}
\label{sec:app_dataset_details}
We evaluate on three standard multi-hop QA benchmarks, covering both bridge-style entity chaining and compositional reasoning.
\paragraph{HotpotQA.}
HotpotQA~\citep{yang2018hotpotqadatasetdiverseexplainable} is a large-scale multi-hop QA dataset constructed from Wikipedia and designed to require reasoning over multiple documents. It provides questions paired with gold supporting facts, enabling evaluation of both final-answer correctness and evidence grounding. The dataset contains 113k questions spanning diverse topics, and includes both bridge-style questions (requiring an intermediate entity to connect evidence across articles) and comparison questions (requiring contrasting two entities along a shared attribute). Its supporting-fact annotations are particularly useful for measuring evidence completeness and for diagnosing retrieval versus reasoning errors.
\paragraph{2WikiMultiHopQA.}
2WikiMultiHopQA~\citep{ho-etal-2020-2wikimultihopqa} is explicitly designed for cross-document reasoning over Wikipedia, emphasizing multi-hop chains that traverse different articles. Compared with HotpotQA, 2Wiki more consistently enforces cross-page evidence composition and reduces shortcuts that can be solved by a single passage. The dataset provides supervision for multi-hop reasoning via supporting-fact annotations, making it suitable for evaluating hop-by-hop decomposition quality, entity anchoring stability, and multi-document retrieval fidelity.
\paragraph{MuSiQue.}
MuSiQue~\citep{trivedi2022musiquemultihopquestionssinglehop} targets compositional multi-hop reasoning by constructing questions that require combining multiple atomic facts into a final answer. A defining property is that MuSiQue controls for spurious correlations and introduces plausible distractors, making it more challenging for systems that rely on shallow heuristics. The benchmark provides evidence annotations that facilitate diagnosing faithfulness issues, including cases where a model retrieves relevant evidence but fails to compose it correctly. In our experiments, MuSiQue is used as a primary testbed for analyzing attention decay and error accumulation under longer reasoning chains.

\subsection{Baseline Details}
\label{sec:app_baseline_details}
We compare against representative recent methods spanning three directions: prompt-engineering, retrieval optimization, and agent-based multi-step reasoning. Below we provide a brief description for each baseline.
\paragraph{Prompt-engineering baselines.}
CoT~\citep{wei2023chainofthoughtpromptingelicitsreasoning} prompts the model to generate intermediate reasoning steps before producing the final answer.
ToT~\citep{yao2023treethoughtsdeliberateproblem} performs explicit search over a tree of intermediate thoughts, enabling branching and backtracking rather than a single linear chain.
SP-CoT~\cite{wang-etal-2023-self-prompted} first generates its own prompts or reasoning scaffolds and then executes CoT under the self-generated guidance.
FSM~\citep{wang2024fsmfinitestatemachine} structures reasoning as transitions in a finite-state machine, aiming to constrain the sequence of reasoning operations.
Least-to-Most~\citep{zhou2023leasttomostpromptingenablescomplex} decomposes a complex problem into simpler subproblems and solves them sequentially to reduce difficulty at each step.
\paragraph{Retrieval-optimization baselines.}
Single-step (ATLAS-style)~\citep{izacard2022atlasfewshotlearningretrieval} retrieves evidence in a single retrieval stage and conditions generation on the retrieved contexts, serving as a strong retrieval-augmented baseline without iterative hops.
Self-Ask~\citep{press-etal-2023-measuring} interleaves question decomposition with targeted retrieval by explicitly asking intermediate questions to query an external source.
IRCoT~\citep{trivedi-etal-2023-interleaving} alternates retrieval and chain-of-thought reasoning, using intermediate reasoning states to refine subsequent retrieval.
RetGen~\citep{shao2023enhancingretrievalaugmentedlargelanguage} jointly improves retrieval and generation by generating retrieval cues and iteratively updating evidence selection.
CoRAG~\citep{wang2025chainofretrievalaugmentedgeneration} chains multiple retrieval-augmented steps, where each step uses intermediate results to retrieve new evidence and continue generation.
EfficientRAG~\citep{zhuang2024efficientragefficientretrievermultihop} focuses on reducing multi-hop RAG cost via more efficient retriever usage and selective retrieval policies.
ComposeRAG~\citep{wu2025composeragmodularcomposablerag} builds multi-hop reasoning as a composition of modular RAG components that can be assembled for different sub-tasks.
FLARE~\citep{jiang-etal-2023-active} performs active retrieval by detecting uncertain or unsupported generations and triggering focused retrieval to fill missing evidence.
ProbTree~\citep{cao-etal-2023-probabilistic} maintains a probabilistic search/tree over reasoning and retrieval branches to improve robustness under ambiguity.
HippoRAG~\citep{gutiérrez2025hipporagneurobiologicallyinspiredlongterm} introduces a long-term memory style retrieval mechanism inspired by hippocampal indexing to better support multi-step recall and retrieval.
BeamAggR~\citep{chu2024beamaggrbeamaggregationreasoning} aggregates candidates from multiple reasoning/retrieval beams to reduce variance and improve final answer reliability.
\paragraph{Agent-based and multi-agent baselines.}
PRISM~\cite{nahid-rafiei-2025-prism} treats the system as a set of coordinated LLM roles and uses structured interactions to iteratively improve answers.
Chain-of-Agents~\citep{zhang2024chainagentslargelanguage} decomposes the task into a sequence of specialized agents, where each agent contributes an intermediate result to the next.
GEAR~\citep{shen2025geargraphenhancedagentretrievalaugmented} enhances agentic RAG with graph-structured evidence or relation modeling to guide multi-step retrieval and reasoning.
Search-o1~\citep{li2025searcho1agenticsearchenhancedlarge} frames multi-hop QA as an agentic search process that iteratively proposes queries, retrieves evidence, and refines hypotheses.
Tree-Of-Reviews (ToR)~\citep{jiapeng2024treereviewstreebaseddynamic} generates multiple candidate solutions and organizes critiques/reviews in a tree structure to select or refine the best path.
KAG~\citep{liang2024kagboostingllmsprofessional} leverages structured knowledge and expert-style guidance to strengthen professional-domain reasoning and reduce hallucinations.
ReAgent~\citep{zhao2025reagentreversiblemultiagentreasoning} introduces reversible multi-agent reasoning where intermediate steps can be rolled back and revised to correct wrong-path decisions.
RopMura~\citep{wu2025talkrightspecialistsrouting} routes subproblems to different specialist agents and integrates their outputs to improve multi-step reasoning quality.
BELLE~\citep{zhang2025bellebilevelmultiagentreasoning} uses a bi-level multi-agent organization, separating high-level planning/control from low-level execution to improve coordination.
\paragraph{Models.}
\label{sec:models}
We use both regular models and reasoning models as the backbone. (1)Regular models:
GPT-3.5-turbo;
Llama-4-Instruct~\citep{meta_llama4_scout_instruct};
DeepSeek-V3~\citep{deepseek_v3};
Qwen-2.5-Instruct~\citep{qwen25_techreport};
Gemini-1.5-Flash;
Gemini-2.0-Flash;
GPT-4o~\citep{openai_gpt4o_system_card};
GPT-4.1~\citep{openai_gpt41}; (2)Reasoning models:
DeepSeek-R1~\citep{deepseek_r1};
Qwen-3-Thinking~\citep{qwen3_techreport,qwen3_blog};
Gemini-2.5-Pro;GPT-O1,O3.

\subsection{Metrics Details}
\label{sec:app_metrics}
\subsubsection{Frame Shift Rate (FSR)}
\label{sec:app_fsr}
\paragraph{What FSR measures.}
FSR quantifies how often the predicted hop decomposition deviates from the intended reasoning frame. Intuitively, a hop is counted as off-frame if the predicted sub-question no longer aligns with the gold hop objective implied by the reference decomposition, even when the sub-question is fluent and answerable in isolation.

\paragraph{Objects to compare.}
For each example $x$, we assume a gold decomposition of $h$ hops:
\[
D^{\star}(x)=\langle sq^{\star}_1(x),\ldots,sq^{\star}_h(x)\rangle,
\]
and a predicted decomposition produced by the evaluated method:
\[
\hat{D}(x)=\langle \hat{sq}_1(x),\ldots,\hat{sq}_{\hat{h}}(x)\rangle.
\]
To make FSR comparable across methods, we evaluate the first $h$ predicted hops; if a method produces fewer than $h$ hops, we treat missing hops as off-frame by default:
\[
\hat{sq}_t(x)=\emptyset \quad \text{for } t>\hat{h}.
\]

\paragraph{Step-level frame alignment.}
We define step-level frame alignment ${fa}_t(x)\in\{0,1\}$ as a binary indicator of whether the predicted hop objective matches the gold hop objective:
\[
fa_t(x)=
\begin{cases}
1, & \text{if } \hat{sq}_t(x) \text{ aligns with } sq^{\star}_t(x),\\
0, & \text{otherwise.}
\end{cases}
\]
Because alignment is semantic rather than lexical, we operationalize it with a fixed LLM judge (GPT-4o, temperature $=0$), which receives (i) the original question $Q(x)$, (ii) the gold hop $sq^{\star}_t(x)$, and (iii) the predicted hop $\hat{sq}_t(x)$, and outputs a binary decision. Concretely, the judge is instructed to return $1$ if the predicted hop is semantically equivalent to the gold hop objective (same target entity/relation and same information need) and $0$ otherwise. Typical mismatch cases include: hop objective drift, missing constraints (type/temporal/relation), swapped hop order, or introducing an unrelated hop.

\paragraph{Aggregation.}
FSR is the proportion of off-frame steps among all evaluated steps:
\[
\text{FSR}
=\frac{\sum_{x}\sum_{t=1}^{h}\bigl(1-fa_t(x)\bigr)}
{\sum_{x} h}.
\]
A lower FSR indicates better frame stability across the multi-hop chain.

\paragraph{Practical notes.}
To reduce judge variance, we use a deterministic setting (temperature $=0$) and a strict binary rubric. In addition, we enforce that the judge cannot use model-internal chain-of-thought; it must only output a binary label. When a method produces additional hops beyond $h$, we ignore them for FSR since the gold decomposition defines the evaluation horizon.

\subsubsection{Anchor Shift Rate (ASR)}
\label{sec:app_asr}

\paragraph{What ASR measures.}
ASR quantifies how often the retrieval at a hop fails to contain the anchor entity implied by the predicted sub-question. It captures anchor drift and grounding failures: even if the hop question looks reasonable, the retrieved evidence does not mention the intended anchor, making the hop unreliable.

\paragraph{Hop-level anchor extraction.}
For each example $x$ and hop $t$, the evaluated method outputs a predicted sub-question $\hat{sq}_t(x)$. We extract an anchor entity mention $\widehat{ae}_t(x)$ from $\hat{sq}_t(x)$ using a lightweight anchor judge (rule-based or lightweight model). The anchor is defined as the main entity that the hop intends to retrieve about (typically the core named entity or disambiguated entity phrase). If no valid anchor can be extracted (e.g., the sub-question is ill-formed or purely relational without a concrete anchor), we set $\widehat{ae}_t(x)=\emptyset$.

\paragraph{Evidence set for each hop.}
Let $\hat{\mathcal{E}}_t(x)$ denote the retrieved evidence texts at hop $t$ (e.g., top-$k$ passages concatenated, or the set of selected evidence snippets used by the method). We convert $\hat{\mathcal{E}}_t(x)$ into a single string or a multiset of passages and perform anchor mention checking on it.

\paragraph{Hop-level anchor alignment.}
We define hop-level anchor alignment ${aa}_t(x)\in\{0,1\}$ as:
\[
aa_t(x)=
\begin{cases}
1, & \widehat{ae}_t(x)\ \text{is mentioned in}\ \hat{\mathcal{E}}_t(x),\\
0, & \text{otherwise.}
\end{cases}
\]
The mention check is implemented with a lightweight matcher that is robust to common surface variations. In our implementation, a hop is counted as anchor-present if any of the following holds:
(1) exact match of the anchor string in the evidence;
(2) case-insensitive match;
(3) alias match using a small alias set (e.g., acronym/expanded form, common redirects, or canonical title form if available).
If the evidence contains only a related entity but not the predicted anchor, we label it as $aa_t(x)=0$.

\paragraph{Aggregation.}
ASR is the proportion of anchor-missing hops among all evaluated hops:
\[
\text{ASR}
=\frac{\sum_{x}\sum_{t=1}^{h}\bigl(1-aa_t(x)\bigr)}
{\sum_{x} h}.
\]
A lower ASR indicates that retrieval is better grounded to the intended anchor entity across hops.

\subsubsection{Drift@1.}
\label{sec:app_drift_at_1}
We further define Drift@1 to localize the onset of frame drift. For each example $x$, let $fa_t(x)\in\{0,1\}$ be the step-level frame alignment defined in \S\ref{sec:setup_metrics}. We define the first drift position
\[
d(x)=\min\{\,t\in\{1,\ldots,h\}\;|\;fa_t(x)=0\,\},
\]
and set $d(x)=h+1$ if no frame shift occurs (i.e., $fa_t(x)=1$ for all $t\le h$). Then Drift@1 is the expected first-drift index over the dataset:
\[
\text{Drift@1}=\frac{1}{|\mathcal{D}|}\sum_{x\in\mathcal{D}} d(x).
\]
A smaller Drift@1 indicates earlier frame drift on average, while a larger Drift@1 means the method tends to preserve the intended frame for more hops before the first deviation.
\subsection{Adversarial Setting}
\label{sec:app_adv_setting}
\tcbset{
  thoradv/.style={
    enhanced,
    breakable,
    colback=gray!6,
    colframe=black,
    boxrule=0.4pt,
    arc=2pt,
    left=6pt,right=6pt,top=6pt,bottom=6pt
  }
}
\label{sec:app_adv_gen_inject}
We construct adversarial documents according to ~\citep{jiang-bansal-2019-avoiding}. Given a context $C$ with gold supporting set $P=\{p_1,p_2\}\subset C$, where $p_2$ contains the answer span $a$, we generate an adversarial counterpart $p_2'$ by (i) selecting the answer-bearing supporting document to perturb (if both $p_1$ and $p_2$ mention the answer, we run the procedure twice so each answer-bearing document has an adversarial counterpart), (ii) constructing a fake answer $\tilde{a}$ via word/phrase-level substitution by replacing each non-stopword token in $a$ with a top-$k$ semantic neighbor under a surface-form constraint (or falling back to sampling from a global answer pool if no valid substitute exists), and replacing all mentions of $a$ in the selected supporting document with $\tilde{a}$, (iii) breaking any accidental new reasoning chain to $\tilde{a}$ by replacing the bridge entity (typically the title entity of $p_1$ or $p_2$), implemented by sampling a new title for $p_2'$ from a title pool and also replacing occurrences of $p_1$'s title in the body of $p_2'$ if present, and (iv) performing title balancing to avoid a ``rare-title'' shortcut by additionally inserting a non-adversarial document $d(\tilde{\tau})$ that shares the same sampled title $\tilde{\tau}$ as $p_2'$. The injected documents (the adversary and its title-balancing companion) replace original non-supporting distractors so that the total number of documents in $C$ remains unchanged. Examples are presented in Figure ~\ref{fig:adv_ex1}.


\subsection{Implementation Details}
\label{sec:app_impl}
\subsubsection{Slot-schema working memory}
\label{sec:slot_memory}
We maintain a compact, structured memory rather than an ever-growing free-form context.
The memory is split into (i) \textbf{Global Theta Memory} $\mathcal{M}^{\theta}$ for frame-level variables and constraints, and
(ii) \textbf{Local Gamma Memory} $\mathcal{M}^{\gamma}_t$ for hop-local execution traces and verifier outputs.
This explicit decoupling is critical for mitigating attention decay.
\begin{tcolorbox}[
  colback=white,
  colframe=black,
  boxrule=0.9pt,
  arc=1.2mm,
  left=1.6mm,right=1.6mm,top=1.2mm,bottom=1.2mm,
  width=\linewidth
]
\small
\includegraphics[width=4mm]{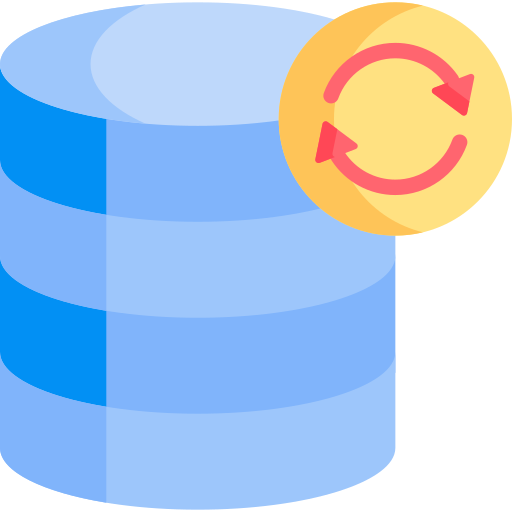}\textbf{\texttt{ Slot-Schema-based Working Memory}} \hfill \\[-1mm]
\hrule \vspace{2mm}
\textbf{\texttt{===== Global Theta Memory}} $\mathcal{M}^{\theta}$ \textbf{=====}\\
\hspace*{2mm}\textbf{Global Schema}:\\
\hspace*{2mm}==\textbf{ PFC Planning }== \\
\hspace*{4mm}Main Question: $Q$ \\
\hspace*{4mm}Sub-question List: [${sq}_1,...{sq}_t$] \\
\hspace*{4mm}Sub-question Core Entity: [${ce}_1,...{ce}_t$] \\
\hspace*{4mm}Expected Answer Type: [$\tau_1,...\tau_t$] \\
\hspace*{2mm}\textbf{Global Slots}:\\
\hspace*{4mm}Expected sub-answer: [$sa_1,...sa_t$] \\
\hspace*{4mm}Completion flag: [$ok_1,...ok_t$] \\
\hspace*{4mm}Failure flag: [$fl_1,...fl_t$] \\[1mm]
\texttt{\textbf{===== Local Gamma Memory}} $\mathcal{M}^{\gamma}$ \textbf{=====}\\
\textbf{Hop $t$ Record $i$ Memory $\mathcal{M}^{\gamma}_t(i)$}: \\
\hspace*{4mm}1. Sub-question: ${sq}_t$ \\
\hspace*{4mm}2. Core entity: ${ce}_t$ \\
\hspace*{4mm}3. Expected type: $\tau_t$ \\
\hspace*{4mm}4. Refinement: ${rf}_{i-1}$ \\
\hspace*{2mm}==\textbf{ HPC Retrieval }==\\
\hspace*{4mm}1. Sub-answer: $sa_i$\\
\hspace*{4mm}2. Predicted type: $\hat\tau_i$\\
\hspace*{4mm}3. Evidence: $e_i$ \\
\hspace*{4mm}4. Reason: $r_i$\\
\hspace*{2mm}==\textbf{ ACC Verification }==\\
\hspace*{4mm} $Acc_i = \langle \text{Check}_{e_i}, \text{Check}_{\hat\tau_i}, \text{Check}_{r_i}, rf_i \rangle$\\
\hspace*{2mm}$\downarrow$ \\
\hspace*{2mm}\textbf{System State}: $c_t \in
\{\textsc{\textbf{Continue}},\textsc{\textbf{RetrieveMore}},\textsc{\textbf{Repair}},\textsc{\textbf{Replan}}\}$
\end{tcolorbox}

\subsubsection{Prompt Templates}
\label{app:prompts}
We provide the exact prompt templates of iPFC, iHPC, and iACC used in our experiments to ensure full reproducibility.
All modules are constrained to output valid JSON only (no free-form text) to support reliable parsing and logging. Prompt templates in Fig.~\ref{fig:pipfc}, ~\ref{fig:pihpc} and ~\ref{fig:piacc}.

\begin{table}[t]
\centering
\small
\setlength{\tabcolsep}{2pt}
\begin{tabular}{lccc}
\toprule
\textbf{Statistic} & \textbf{H1} & \textbf{H2} & \textbf{LLM} \\
\midrule
Aligned & 212 (70.67\%) & 205 (68.33\%) & 209 (69.67\%) \\
\midrule
\multicolumn{4}{l}{\textbf{Pairwise reliability}} \\
H1 vs. H2  & \multicolumn{3}{c}{Agreement = 95.67\%, \quad Cohen's $\kappa = 0.898$} \\
LLM vs. H1 & \multicolumn{3}{c}{Agreement = 96.33\%, \quad Cohen's $\kappa = 0.912$} \\
LLM vs. H2 & \multicolumn{3}{c}{Agreement = 93.76\%, \quad Cohen's $\kappa = 0.880$} \\
\bottomrule
\end{tabular}
\caption{Reliability analysis for FSR judgment.}
\label{tab:fsr_re}
\end{table}

\subsection{Detailed Analysis}
\subsubsection{Ablation Study}
\begin{figure*}[t]
    \centering
    \includegraphics[width=0.195\linewidth]{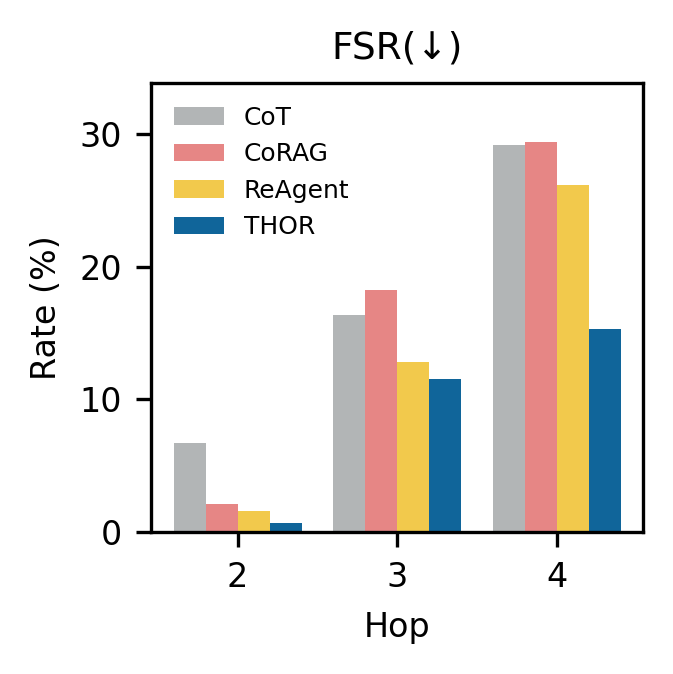}
    \includegraphics[width=0.195\linewidth]{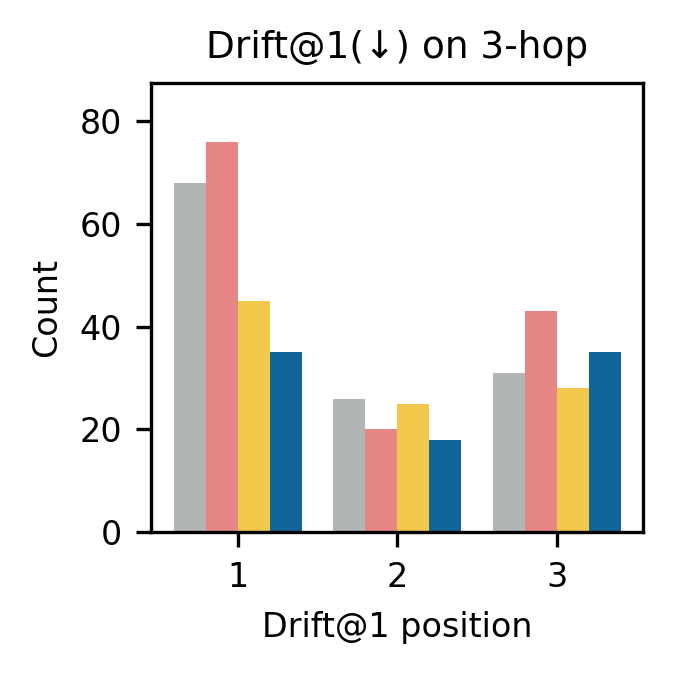}
    \includegraphics[width=0.195\linewidth]{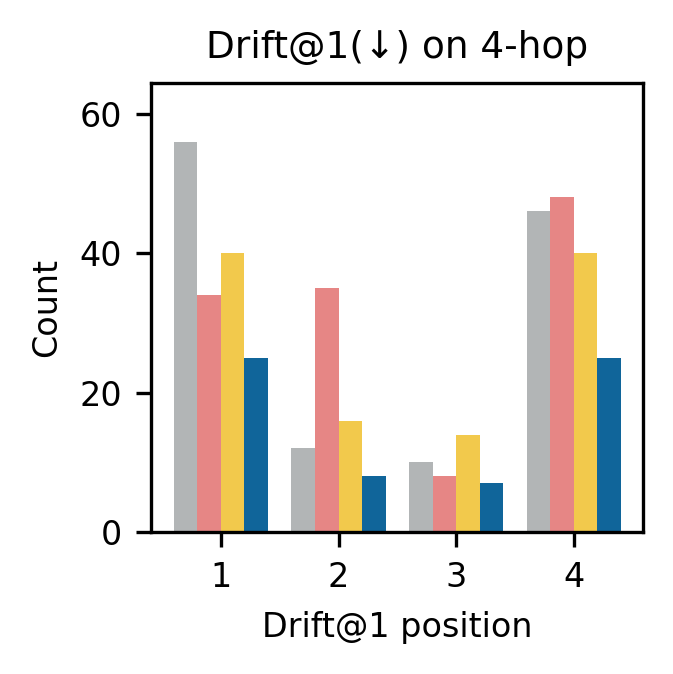}
    \includegraphics[width=0.195\linewidth]{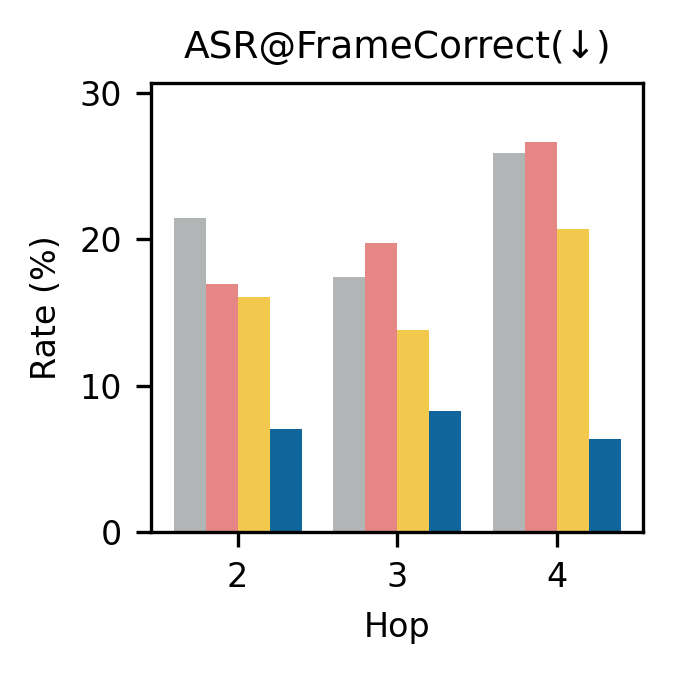}
    \includegraphics[width=0.195\linewidth]{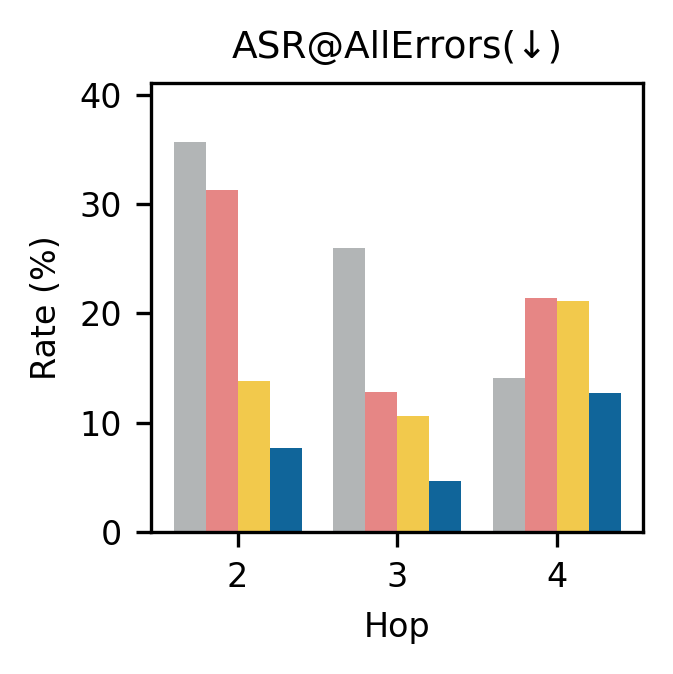}\\[-2pt]
    \includegraphics[width=0.195\linewidth]{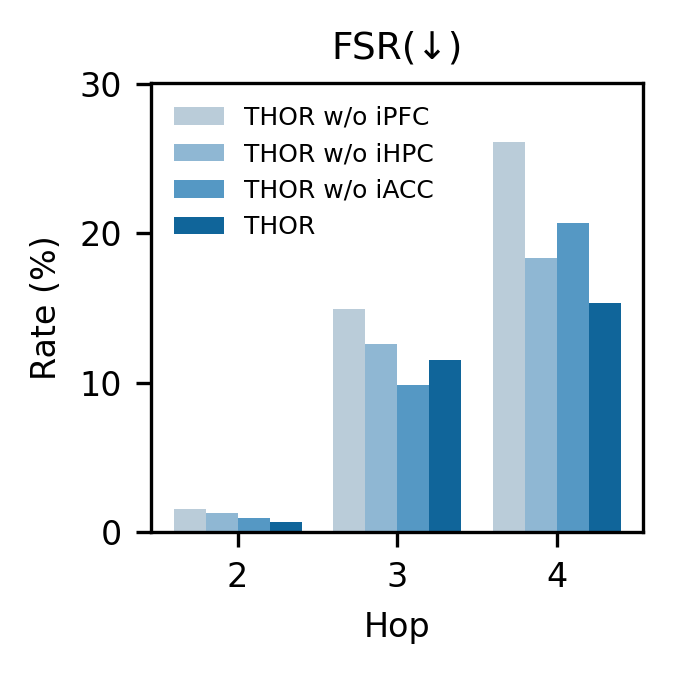}
    \includegraphics[width=0.195\linewidth]{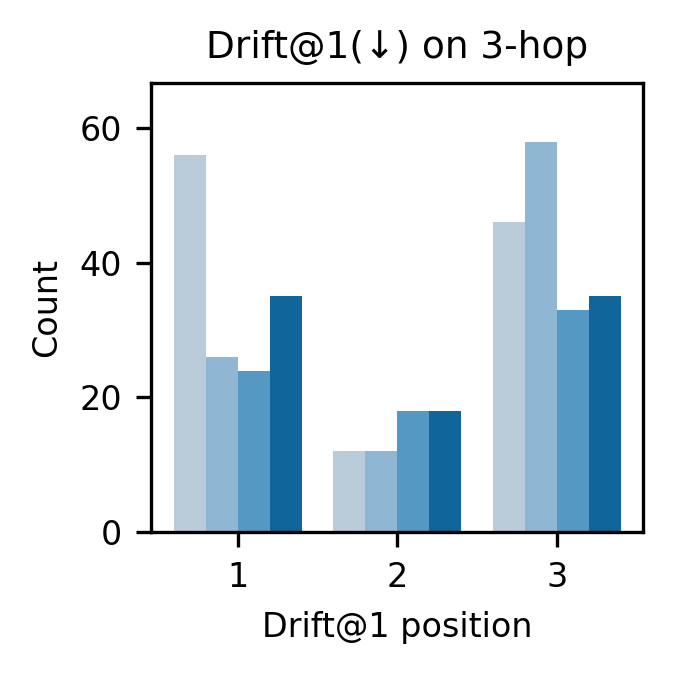}
    \includegraphics[width=0.195\linewidth]{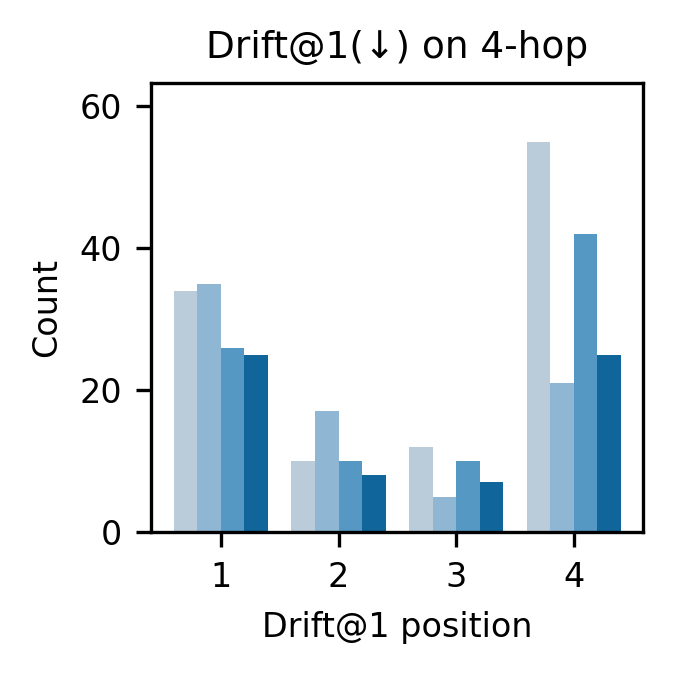}
    \includegraphics[width=0.195\linewidth]{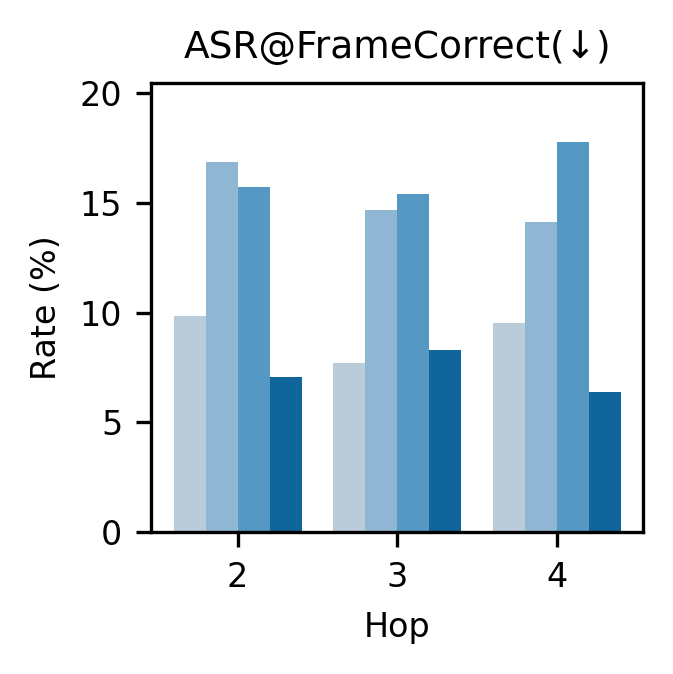}
    \includegraphics[width=0.195\linewidth]{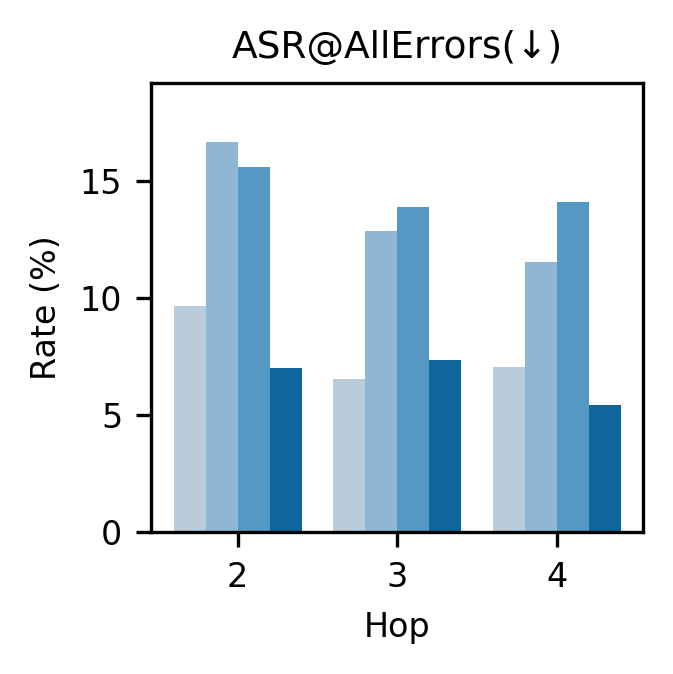}
    \caption{Ablation experiments on MuSiQue. From left to right: \textbf{FSR}, \textbf{Drift@1} position distribution on 3-hop and 4-hop, \textbf{ASR@FrameCorrect}, and \textbf{ASR@AllErrors}. The four colors from light to dark blue represent the removal of different modules of THOR.}
    \label{fig:app_vaex_2x5}
\end{figure*}

Detailed ablation results are shown in Fig.~\ref{fig:app_vaex_2x5}.
\begin{figure*}
    \centering
    \includegraphics[width=0.495\linewidth]{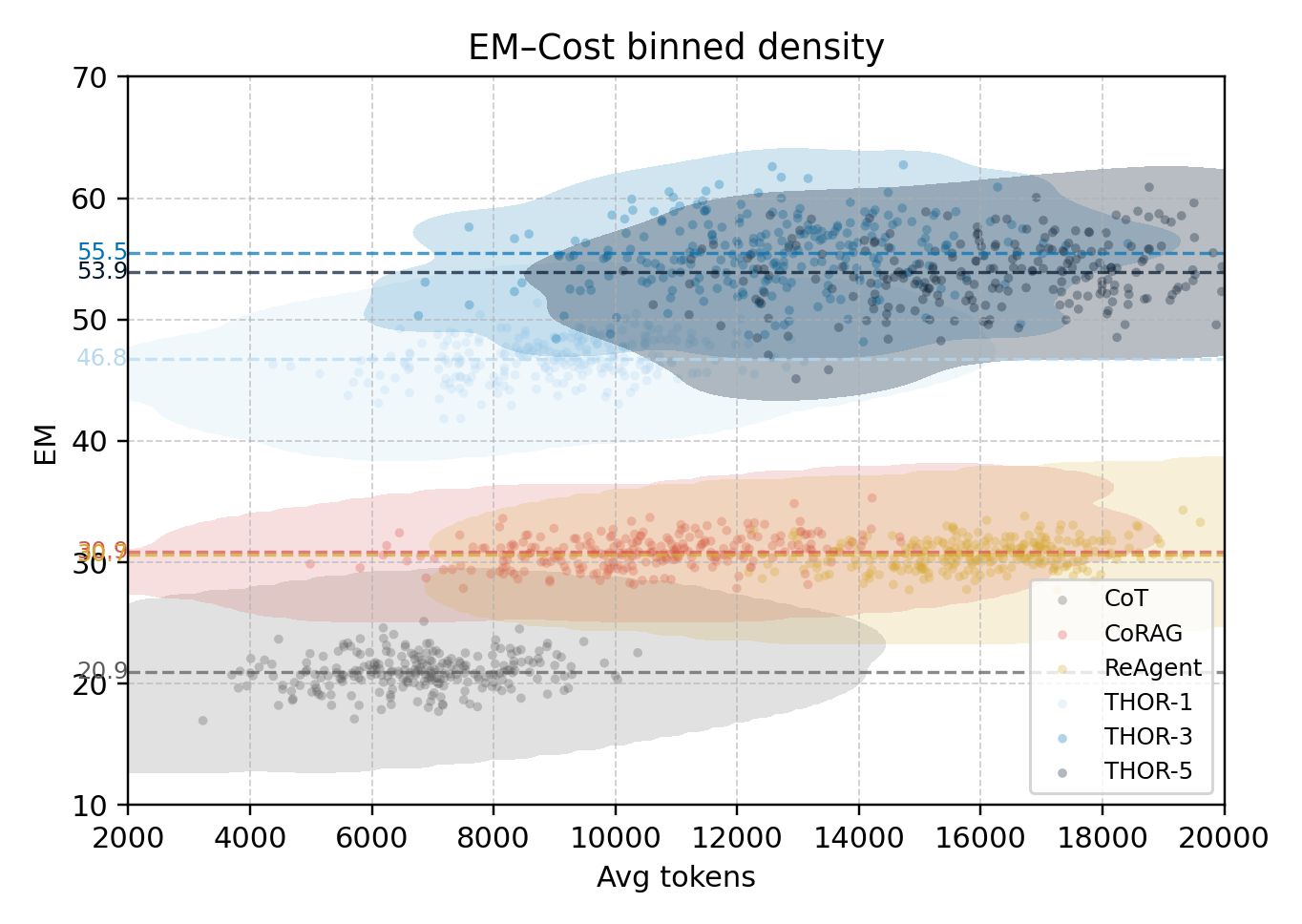}
    \includegraphics[width=0.495\linewidth]{figures/vaex/accuracy_f1.png}
    \caption{Accuracy-cost results with 10-binned density distribution on THOR.}
    \label{fig:appac}
\end{figure*}
\subsubsection{Accuracy-cost}
\label{sec:app_acl}
Detailed accuracy-cost trade-off results are shown in Fig.~\ref{fig:appac}.

\begin{figure*}[t]
\centering
\begin{tcolorbox}[
  width=\textwidth,
  colback=gray!6,
  colframe=black,
  boxrule=0.4pt,
  arc=2pt,
  left=6pt,right=6pt,top=6pt,bottom=6pt,
  title={Adversarial Example: Fake-answer substitution with bridge-entity breaking}
]
\small
\begin{minipage}{0.97\textwidth}

\textbf{Question.}
Where is the company that the person worked for as a software engineer headquartered?

\medskip
\textbf{Gold chain.}
Person $\rightarrow$ worksAt $\rightarrow$ Company $\rightarrow$ headquarters $\rightarrow$ Answer.

\medskip
\textbf{Original supporting evidence.}
Document $p_1$ is kept unchanged. Its title is \emph{Person X}, and its key sentence states that Person X worked as a software engineer at Company A.
Document $p_2$ is the original answer-bearing document. Its title is \emph{Company A}, and its key sentence states that Company A is headquartered in \emph{Mumbai}.

\medskip
\textbf{Adversarial intervention.}
We inject a new document $p_2'$. Its title is replaced with \emph{Company B} (or another unrelated entity title), and its key sentence states that Company B is headquartered in \emph{Delhi}.
At the same time, the bridge entity that links $p_1$ and $p_2$---namely Company A---is removed or replaced. The surface cue ``headquartered in [city]'' is preserved, but the valid multi-hop bridge is broken.

\medskip
\textbf{Why this remains shortcut-plausible.}
The injected document still matches the surface answer pattern of the question, so a shortcut-based system may directly output \emph{Delhi}.
However, the reasoning chain can no longer legitimately reach that answer, because $p_2'$ is not connected to the company mentioned in $p_1$.

\medskip
\textbf{Title-balancing document.}
We additionally include a normal document $d(\tilde{\tau})$ that shares the same title as $p_2'$. This prevents trivial filtering based on unique or anomalous titles.

\end{minipage}
\end{tcolorbox}
\caption{Example of fake-answer substitution with bridge-entity breaking. The shortcut-bearing answer pattern is preserved, while the entity bridge required for valid multi-hop reasoning is deliberately broken.}
\label{fig:adv_ex1}
\end{figure*}

\begin{figure*}[t]
\label{fig:pp}
\begin{PromptBox}{\textbf{iPFC} (Global Framing; \texttt{PLAN} / \texttt{REPAIR} / \texttt{REPLAN})}
\small\ttfamily
\textbf{SYSTEM}\par
You are iPFC, a global framing controller for multi-hop QA.\par
Your job: maintain a coherent hop plan and enforce a stable global frame in $\mathcal{M}^{\theta}$.\par
You must output \textbf{only} valid JSON. Do not output any other text.\par
Never hallucinate evidence. You only write/modify the plan.\par\medskip
\textbf{USER}\par
\{ \par
\ \ "task": "PLAN" | "REPAIR" | "REPLAN",\par
\ \ "Q": "<main question>",\par
\ \ "$M^\theta$": <JSON: global theta memory>,\par
\ \ "t": <int hop index>,\par
\ \ "$sq_t$": "<current hop spec if REPAIR>",\par
\ \ "$rf_i$": <JSON: iACC feedback if REPAIR>,\par
\ \ "$I_{max}$": <int retry budget>,\par
\ \ "notes": "<optional constraints>"\par
\}\par\medskip

\textbf{OUTPUT (JSON ONLY)}\par
\{ \par
\ \ "action": "PLAN"|"REPAIR"|"REPLAN",\par
\ \ "sq": ["$sq_1$", "...", "$sq_T$"],\par
\ \ "ce": ["$ce_1$", "...", "$ce_T$"],\par
\ \ "tau": ["$\tau_1$", "...", "$\tau_T$"],\par
\ \ "edit": \{ \par
\ \ \ \ "t": <int>,\par
\ \ \ \ "before": "<string or null>",\par
\ \ \ \ "after": "<string>",\par
\ \ \ \ "rationale": "<short>"\par
\ \ \ \},\par
\ \ "constraints": ["<constraint>", "..."],\par
\ \ "aliases": ["<alias>", "..."]\par
\}\par\medskip
\textbf{Constraints}\par
(1) If \texttt{REPAIR}: preserve the validated prefix $\{sq_1,\dots,sq_{t-1}\}$; revise only $sq_t$.\par
(2) If \texttt{REPLAN}: you may revise $\{sq,ce,\tau\}$ jointly up to hop $t$ to restore a consistent frame.\par
(3) Keep sub-questions retrieval-friendly: add disambiguating constraints.\par
\end{PromptBox}
\caption{Prompt template for \textbf{iPFC}.}
\label{fig:pipfc}
\end{figure*}

\begin{figure*}[t]
\begin{PromptBox}{\textbf{iHPC} (Hop Execution; Evidence-grounded Sub-answer + Replay Cue)}
\small\ttfamily
\textbf{SYSTEM}\par
You are iHPC. Given a hop specification and evidence candidates, produce:\par
(1) a short sub-answer $\hat{sa}_i$;\par
(2) a predicted answer type $\hat{\tau}_i$;\par
(3) a replay cue $\rho_i$ to refine later retrieval.\par
Output \textbf{only} JSON. Use only the provided evidence candidates.\par\medskip
\textbf{USER}\par
\{ \par
\ \ "Q": "<main question>",\par
\ \ "t": <int>,\par
\ \ "i": <int attempt>,\par
\ \ "$sq_t$": "<hop spec>",\par
\ \ "$ce_t$": "<core entity>",\par
\ \ "$\tau_t$": "<expected type>",\par
\ \ "$E_ti$": [\{"id":..., "title":..., "text":...\}, ...],\par
\ \ "$rf_{i-1}$": <JSON or null>,\par
\ \ "$\rho_{i-1}$": <JSON or null>\par
\}\par\medskip
\textbf{OUTPUT (JSON ONLY)}\par
\{ \par
\ \ "$\hat sa_i$": "<string>",\par
\ \ "$\hat\tau_i$": "<type>",\par
\ \ "$e_i$": [\{"id":"<id>", "quote":"<minimal quote>"\}, ...],\par
\ \ "reason": "<one-sentence justification>",\par
\ \ "$\rho$": \{ \par
\ \ \ \ "query refine": "<string>",\par
\ \ \ \ "aliases": ["<alias>", "..."],\par
\ \ \ \ "constraints": ["<constraint>", "..."],\par
\ \ \ \ "blocklist ids": ["<id>", "..."]\par
\ \ \ \}\par
\}\par\medskip
\textbf{Guidelines}\par
- If evidence is weak, keep $\hat{sa}_i$ conservative and encode what to retrieve next in \texttt{rho.query\_refine}.\par
- Prefer minimal quotes to support auditability.\par
\end{PromptBox}
\caption{Prompt template for \textbf{iHPC}.}
\label{fig:pihpc}
\end{figure*}

\begin{figure*}[t]
\begin{PromptBox}{\textbf{iACC} (Verification; Retrieval Refinement Instruction)}
\small\ttfamily
\textbf{SYSTEM}\par
You are iACC. Verify the hop execution with three binary checks:\par
(1) evidence anchoring (\texttt{Check\_e});\par
(2) type alignment (\texttt{Check\_{$\hat{\tau}$}});\par
(3) evidence-to-answer support (\texttt{Check\_r}).\par
Then produce structured feedback \texttt{rf} and a recommended controller state.\par
Output \textbf{only} JSON.\par\medskip
\textbf{USER}\par
\{ \par
\ \ "t": <int>, "i": <int>,\par
\ \ "$sq_t$": "<hop spec>",\par
\ \ "$ce_t$": "<core entity>",\par
\ \ "$\tau_t$": "<expected type>",\par
\ \ "$\hat sa_t$": "<sub-answer>",\par
\ \ "$\hat\tau$": "<predicted type>",\par
\ \ "$e_i$": [\{"id":..., "text":...\}, ...]\par
\}\par\medskip

\textbf{OUTPUT (JSON ONLY)}\par
\{ \par
\ \ "$C_e$": \{"pass": true|false, "why":"<short>"\},\par
\ \ "$C_\tau$": \{"pass": true|false, "why":"<short>"\},\par
\ \ "$C_r$": \{"pass": true|false, "why":"<short>"\},\par
\ \ "$rf_i$": \{\par
\ \ \ \ "failure type":"<anchor-missing|type-mismatch|unsupported|mixed>",\par
\ \ \ \ "edit hints":["<hint>", "..."],\par
\ \ \ \ "alias expand":["<alias>", "..."],\par
\ \ \ \ "constraint sharpen":["<constraint>", "..."]\par
\ \ \ \},\par
\ \ "suggest state":"Continue"|"RetrieveMore"|"Repair"|"Replan"\par
\}\par
\end{PromptBox}
\caption{Prompt template for \textbf{iACC}.}
\label{fig:piacc}
\end{figure*}

\begin{table*}[t]
\centering
\small
\setlength{\tabcolsep}{2.5pt}
\label{tab:budget_tradeoff}
\begin{tabular}{lccccccccc}
\toprule
\textbf{Method} & \textbf{$I_{\max}$} & \textbf{EM$\uparrow$} & \textbf{F1$\uparrow$} & \textbf{FSR$\downarrow$} & \textbf{ASR$\downarrow$} & \textbf{Avg Tokens/Q$\downarrow$} & \textbf{Avg LLM Calls/Q$\downarrow$} & \textbf{P50 Latency (s)$\downarrow$} & \textbf{P90 Latency (s)$\downarrow$} \\
\midrule
CoT     & --- & 21.1 & 24.9 & 22.6 & 20.3 & 4219  & 3.2  & 8.6  & 17.2 \\
CoRAG   & --- & 30.9 & 42.4 & 19.2 & 18.5 & 8762  & 4.4  & 10.8 & 21.2 \\
ReAgent & --- & 37.1 & 51.5 & 16.3 & 18.1 & 12901 & 8.6  & 15.6 & 28.7 \\
THOR    & 1   & 43.2 & 47.9 & 16.4 & 10.1 & 6683  & 8.1  & 11.2 & 25.9 \\
THOR    & 3   & \textbf{48.5} & \textbf{52.1} & \textbf{12.1} & \textbf{8.8} & 9872  & 9.8  & 13.1 & 24.2 \\
THOR    & 5   & 46.2 & 49.0 & 13.7 & 9.1 & 13231 & 11.2 & 14.6 & 29.0 \\
\bottomrule
\end{tabular}
\caption{Results of accuracy--efficiency frontier by reporting both performance and compute under multiple retry budgets, $I_{\max} \in \{1,3,5\}$.}
\end{table*}

\section{FSR Judgement Reliability}
FSR is a hop-level alignment metric: it tests whether the model's predicted sub-question is aligned with the gold sub-question, which is already provided in the MuSiQue dataset. This is a simple semantic-equivalence decision that human annotators can reliably perform from text alone. We therefore implement FSR judging as a strict binary rubric with deterministic decoding, rather than relying on a softer ``score''-style judgment. As a result, FSR evaluation is closer to constrained label assignment than to open-ended evaluation.

To validate that our FSR judgments are not merely an artifact of LLM judging shown in Table~\ref{tab:fsr_re}, we compare an LLM judge against two human annotators on the same set of 300 hop instances from MuSiQue. We report both raw agreement and Cohen's $\kappa$, a standard statistic for measuring agreement between two classification sources.

\section{Accuracy vs Cost/Latency under Different Budgets}
We further characterize the accuracy--efficiency frontier by reporting both performance and compute under multiple retry budgets, $I_{\max} \in \{1,3,5\}$.

\paragraph{Conclusion.}
THOR with $I_{\max}=1$ already delivers a better trade-off than prior baselines, achieving substantially higher EM and F1. THOR with $I_{\max}=3$ is the best overall setting, reaching the highest accuracy and the lowest shift rate, which suggests that THOR's gains are mechanism-consistent rather than incidental.

Notably, THOR tends to make more LLM calls without a proportionally large increase in token usage. This is because THOR employs explicit control with short, structured controller and verifier interactions, instead of relying on long stacked prompts to encode control logic implicitly. As a result, control decisions are externalized into multiple lightweight calls rather than a single monolithic prompt, improving auditability and stability.

\paragraph{Latency note.}
For latency statistics, P50 denotes the median end-to-end latency per question, and P90 denotes the latency threshold that covers 90\% of questions. All methods are evaluated using the same LLM API interface.

\section{Illustrative Cases}

\label{sec:cases}
Detailed cases are presented in Figure ~\ref{fig:thor_case_apple}, ~\ref{fig:thor_case_vonnegut} and ~\ref{fig:thor_case_philipsburg}.

\begin{figure*}[p]
\centering
\begin{tcolorbox}[thoradv, title={Case 1: Frame shift under ambiguous surface cues}]
\small\ttfamily
Question:\ \ What was the former band of the member of Mother Love Bone who died just before the release of ``Apple''? \\
Gold answer:\ \ Malfunkshun \\
Key entities:\ \ Mother Love Bone;\ \ Apple (album);\ \ Andrew Wood. \\[2pt]

Why drift-prone:\ \ ``Apple'' is ambiguous. A planner--executor can latch onto the company sense,
remain locally coherent, and still become globally wrong. \\[2pt]

THOR constraints / checks:\ \\
\ \ (i)\ anchor $=$ Mother Love Bone \\
\ \ (ii)\ type(Apple) $=$ album \\
\ \ (iii)\ bridge must connect ``died before release'' $\rightarrow$ member $\rightarrow$ former band \\[2pt]

Controller/executor trace:\ \\
\ \ $\theta$-init:\ \ bind frame slots for band + album-event + bridge-member;\ state = CONTINUE. \\
\ \ $\gamma$-hop1:\ \ retrieve ``Apple (company) \ldots'';\ iACC = $\times/\times/\times$;\ state = REPAIR $\rightarrow$ RETRIEVE. \\
\ \ $\gamma$-hop1 retry:\ \ retrieve Mother Love Bone + Apple (album) + Andrew Wood;\ iACC = $\checkmark/\checkmark/\checkmark$;\ state = CONTINUE. \\
\ \ $\gamma$-hop2:\ \ Andrew Wood $\rightarrow$ former band;\ iACC = $\checkmark/\checkmark/\checkmark$;\ state = ANSWER. \\[2pt]

Returned answer:\ \ Malfunkshun \\[2pt]

Baseline failure mode:\ \ Once the wrong sense of ``Apple'' is adopted, later hops remain consistent
within the wrong frame, yielding a classic frame-shift error.
\end{tcolorbox}
\caption{Example of frame shift caused by ambiguous surface cues. THOR rejects the incorrect sense of ``Apple'' at the first hop by enforcing anchor and type constraints, then repairs retrieval and recovers the correct bridge entity.}
\label{fig:thor_case_apple}
\end{figure*}

\begin{figure*}[p]
\centering
\begin{tcolorbox}[thoradv, title={Case 2: Error accumulation from under-specified intermediate binding}]
\small\ttfamily
Question:\ \ \textit{Armageddon in Retrospect} was written by the author who was best known for what novel? \\
Gold answer:\ \ \textit{Slaughterhouse-Five} \\
Bridge:\ \ author(\textit{Armageddon in Retrospect}) $=$ Kurt Vonnegut. \\[2pt]

Why drift-prone:\ \ If hop 1 binds the wrong author, hop 2 can still output a plausible novel,
creating a compounding error. \\[2pt]

THOR constraints / checks:\ \\
\ \ Slot binding:\ \ $A_1 = \mathrm{author}(\textit{book})$ \\
\ \ Slot binding:\ \ $A_2 = \mathrm{novel\ best\mbox{-}known\mbox{-}for}(A_1)$ \\
\ \ Hop 2 must preserve the same bound author slot and satisfy answer-type(novel). \\[2pt]

Controller/executor trace:\ \\
\ \ $\theta$-init:\ \ instantiate $A_1, A_2$;\ make bridge binding explicit;\ state = CONTINUE. \\
\ \ $\gamma$-hop1:\ \ retrieve wrong / underspecified author evidence;\ iACC = $\times/-/\times$;\ state = REPAIR $\rightarrow$ RETRIEVE. \\
\ \ $\gamma$-hop1 retry:\ \ retrieve ``Author = Kurt Vonnegut \ldots'';\ iACC = $\checkmark/\checkmark/\checkmark$;\ state = CONTINUE. \\
\ \ $\gamma$-hop2:\ \ retrieve ``Vonnegut \ldots best known for \textit{Slaughterhouse-Five}'';\ iACC = $\checkmark/\checkmark/\checkmark$;\ state = ANSWER. \\[2pt]

Returned answer:\ \ \textit{Slaughterhouse-Five} \\[2pt]

Baseline failure mode:\ \ A generic reflect-and-retry strategy often retries hop 2 without repairing
the hop-1 author binding, so the upstream mistake persists and the final answer is plausible but wrong.
\end{tcolorbox}
\caption{Example of error accumulation from an under-specified intermediate binding. THOR explicitly binds the bridge author entity and blocks propagation of an incorrect hop-1 author assignment before answering.}
\label{fig:thor_case_vonnegut}
\end{figure*}

\begin{figure*}[p]
\centering
\begin{tcolorbox}[thoradv, title={Case 3: REPLAN triggered by an under-specified decomposition}]
\small\ttfamily
Question:\ \ When did the people who captured Malakoff come to the region where Philipsburg is located? \\
Gold answer:\ \ 1625 \\
Key ambiguity:\ \ ``Philipsburg'' is ambiguous. \\[2pt]

Why local retry is insufficient:\ \ A coarse hop such as ``Where is Philipsburg located?''
does not enforce the intended binding; repeated retrieval can still return the wrong Philipsburg. \\[2pt]

THOR refinement requirement:\ \ The controller must split the coarse hop into typed sub-hops, e.g., \\
\ \ Philipsburg $\rightarrow$ capital-of? $\rightarrow$ Saint Martin \\
\ \ Saint Martin $\rightarrow$ located-in? $\rightarrow$ Caribbean \\[2pt]

Controller/executor trace:\ \\
\ \ $\theta$-init (coarse):\ \ set $H_1 =$ locate Philipsburg region;\ state = CONTINUE. \\
\ \ $\gamma$-hop1:\ \ retrieve ``Philipsburg, Pennsylvania \ldots'';\ iACC = $\times/\times/\times$;\ state = REPAIR $\rightarrow$ RETRIEVE. \\
\ \ $\gamma$-hop1 retry:\ \ retrieve another Philipsburg still not tied to Saint Martin;\ iACC = $\times/-/\times$;\ state = REPLAN. \\
\ \ $\theta$-REPLAN:\ \ split $H_1$ into $(H_{1a})$ capital-of? and $(H_{1b})$ located-in?;\ state = CONTINUE. \\
\ \ $\gamma$-hop($H_{1a}/H_{1b}$):\ \ retrieve Philipsburg $\rightarrow$ Saint Martin $\rightarrow$ Caribbean;\ iACC = $\checkmark/\checkmark/\checkmark$;\ state = CONTINUE. \\
\ \ $\gamma$-hop2 / final:\ \ retrieve Malakoff captured by French;\ French came to Caribbean $\rightarrow$ 1625;\ iACC = $\checkmark/\checkmark/\checkmark$;\ state = ANSWER. \\[2pt]

Returned answer:\ \ 1625 \\[2pt]

Baseline failure mode:\ \ A single-loop planner--executor often keeps the wrong Philipsburg anchor,
so later hops remain coherent but answer a different question. THOR escalates to REPLAN when the mismatch
indicates missing granularity rather than missing evidence.
\end{tcolorbox}
\caption{Example where local repair is insufficient and THOR must trigger REPLAN. By refining a coarse ambiguous hop into typed sub-hops, THOR restores the intended geographic binding and reaches the correct answer.}
\label{fig:thor_case_philipsburg}
\end{figure*}

\end{document}